%% file: main.tex
\documentclass{article}

\PassOptionsToPackage{numbers,sort&compress}{natbib}
\usepackage[main,final]{neurips_2026}

\usepackage[utf8]{inputenc}
\usepackage[T1]{fontenc}
\usepackage[hyperfootnotes=false]{hyperref}
\hypersetup{
  pdfborder={0 0 0},
  pdftitle={ClusterSplat: Semantic Cluster Selection for 3D Visual Grounding in Gaussian Splatting},
  pdfauthor={Jaesung Lee, Hyeontaek Hwang, Gabriel Manalu, Daeyoung Kim}
}
\usepackage{url}
\usepackage{microtype}
\usepackage{graphicx}
\usepackage{subcaption}
\usepackage{booktabs}
\usepackage{xcolor}
\usepackage{amsmath}
\usepackage{amssymb}
\usepackage{amsfonts}
\usepackage{mathtools}
\usepackage{amsthm}
\usepackage{multirow}
\usepackage{float}
\usepackage[capitalize,noabbrev]{cleveref}

\ifdefined\pdfsuppresswarningpagegroup
\fi

\theoremstyle{plain}

\theoremstyle{definition}

\theoremstyle{remark}

\newlength{\appendixablationwidth}
\newlength{\maintablewidth}
\colorlet{commentred}{black}
\colorlet{citationblue}{black}
\colorlet{methodblue}{black}
\colorlet{reviewblue}{black}

\colorlet{resultred}{reviewblue}

\colorlet{red}{black}
\colorlet{blue}{black}
\colorlet{editred}{reviewblue}
\definecolor{scorerred}{RGB}{0,0,180}
\definecolor{ablationred}{RGB}{0,0,180}
\definecolor{c0507red}{RGB}{180,0,0}
\definecolor{checklistyesblue}{RGB}{0,0,255}
\definecolor{checklistnoorange}{RGB}{255,128,0}
\definecolor{checklistnagray}{gray}{0.5}
\renewcommand{\answerYes}[1][]{\textcolor{checklistyesblue}{[Yes]#1}}
\renewcommand{\answerNo}[1][]{\textcolor{checklistnoorange}{[No]#1}}
\renewcommand{\answerNA}[1][]{\textcolor{checklistnagray}{[N/A]#1}}

\newif\ifpionereview
\pionereviewfalse
\definecolor{pionerevisioncolor}{RGB}{200,0,0}
\DeclareRobustCommand{\pionerevision}[1]{\ifpionereview\textcolor{pionerevisioncolor}{#1}\else#1\fi}
\definecolor{clusteringcomparisonblue}{RGB}{0,0,200}

\title{ClusterSplat: Semantic Cluster Selection\\
for 3D Visual Grounding in Gaussian Splatting}

\author{
  \pionerevision{Jaesung Lee \quad Hyeontaek Hwang \quad Gabriel Manalu \quad Daeyoung Kim} \\[2pt]
  \pionerevision{School of Computing, KAIST} \\
  \pionerevision{\texttt{\{ljs6017, hwanght, gabrieldavin23, kimd\}@kaist.ac.kr}}
}

\begin{document}

\maketitle

\input{figures/1_teaser.tex}

\input{sections/0_abstract.tex}

\input{sections/1_intro.tex}

\section{Related Work}
\label{sec:related}

\input{sections/2_1_gaussians.tex}
\input{figures/2_overview.tex}
\input{sections/2_2_grounding.tex}

\section{Method}
\label{sec:method}

\input{sections/3_1_problem.tex}
\input{sections/3_2_clustering.tex}
\input{sections/3_3_scorer.tex}
\input{sections/3_4_rendering.tex}

\section{Experiments}
\label{sec:experiments}

\input{tables/1_benchmarks.tex}
\input{figures/3_scanrefer.tex}
\input{sections/4_1_setup.tex}
\input{sections/4_2_results.tex}
\input{sections/4_3_analysis.tex}
\input{sections/5_discussion.tex}
\input{sections/6_conclusion.tex}

\clearpage

\input{sections/7_ack.tex}

{\small
\bibliographystyle{unsrtnat}
\bibliography{main}
}

\appendix
\onecolumn

{
\hypersetup{linkcolor=black}

\noindent\hyperref[sec:implementation]{\textbf{A \ \ \ Implementation Details}}\\
\hspace*{6mm} \hyperref[subsec:system_setup]{A.1 \ \ System Setup and Hyperparameters} \\
\hspace*{6mm} \hyperref[subsec:mask_feature_details]{A.2 \ \ Instance-Aware Gaussian Feature Learning Details} \\
\hspace*{6mm} \hyperref[subsec:adaptive_clustering_details]{A.3 \ \ Scene-Adaptive Clustering and Mask Rendering Details} \\
\hspace*{6mm} \hyperref[subsec:data_prep]{A.4 \ \ Data Preparation} \\
\hspace*{6mm} \hyperref[subsec:auto_supervision]{A.5 \ \ Automated Pseudo Label Pipeline} \\
\hspace*{6mm} \hyperref[subsec:baseline_impl]{A.6 \ \ Baseline Implementation Details} \\[3mm]

\noindent\hyperref[sec:additional_results]{\textbf{B \ \ \ Additional Results}} \\
\hspace*{6mm} \hyperref[subsec:sensitivity]{B.1 \ \ Sensitivity Analysis of Selection Heuristics} \\
\hspace*{6mm} \hyperref[subsec:ablation]{B.2 \ \ Design Choices} \\
\hspace*{6mm} \hyperref[subsec:cluster_quality]{\pionerevision{B.3 \ \ Cluster Evaluation Protocols}} \\
\hspace*{6mm} \hyperref[subsec:caption_sensitivity]{\pionerevision{B.4 \ \ Caption Source and Evaluation-Language Analysis}} \\
\hspace*{6mm} \hyperref[subsec:reflerf_results]{\pionerevision{B.5 \ \ Ref-LERF Quantitative Results}} \\
\hspace*{6mm} \hyperref[sec:efficiency_analysis]{\pionerevision{B.6 \ \ Efficiency Analysis}} \\
\hspace*{6mm} \hyperref[subsec:qualitative]{\pionerevision{B.7} \ \ Additional Qualitative Results} \\
}
\noindent\rule{\linewidth}{0.2pt}

\input{appendix/0_intro.tex}
\section{Implementation Details}
\label{sec:implementation}

\input{appendix/A_1_setup.tex}
\input{appendix/A_2_features.tex}
\input{appendix/A_3_clustering.tex}
\input{appendix/A_4_5_data.tex}
\input{appendix/A_6_baselines.tex}
\input{appendix/B_0_results.tex}
\input{appendix/B_1_sensitivity.tex}

\input{appendix/B_2_design.tex}

\input{appendix/B_3_clusters.tex}
\input{appendix/B_4_captions.tex}
\input{appendix/B_5_reflerf.tex}

\clearpage
\input{appendix/B_6_efficiency.tex}

\clearpage
\input{appendix/B_7_qualitative.tex}

\clearpage
\input{checklist.tex}

\end{document}

%% file: figures/1_teaser.tex
\begin{figure}[H]
    \centering
    \includegraphics[width=0.88\linewidth]{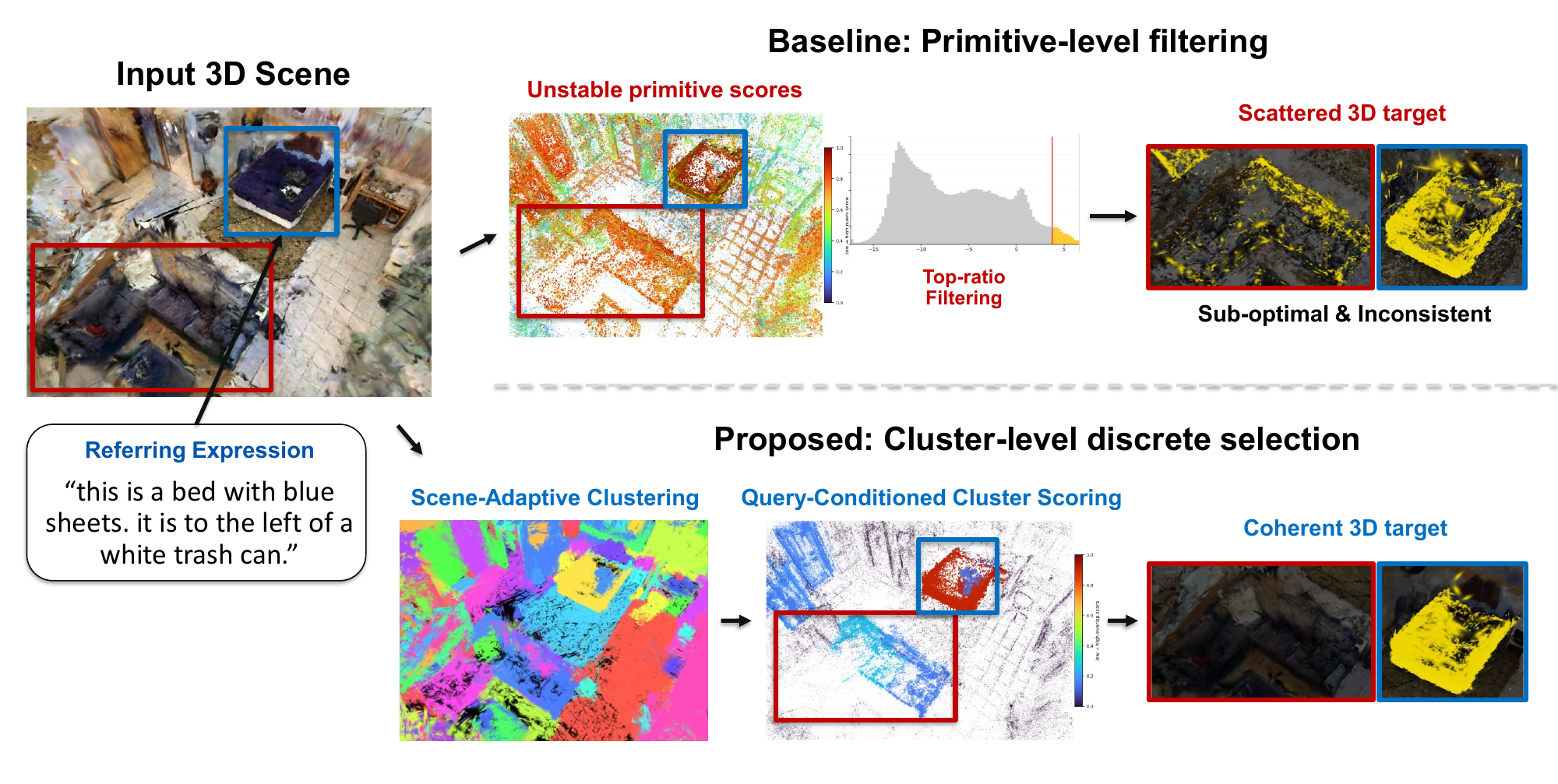}
    \vspace{-1.5mm}
    \caption{\textbf{Primitive-level filtering versus cluster-level selection.} Given a referring expression, primitive score fields can be unstable, and top-ratio filtering may scatter the 3D support across the target and nearby distractors. ClusterSplat instead performs scene-adaptive clustering followed by query-conditioned cluster scoring to select one coherent 3D target.}
    \label{fig:teaser}
\end{figure}
\vspace{-6mm}

%% file: sections/0_abstract.tex
\begin{abstract}
We study 3D visual grounding in 3D Gaussian Splatting (3DGS), where a referring expression should identify a target object as both a rendered 2D mask and a set of explicit 3D Gaussians.
Prior methods typically extract 3D targets by applying heuristic thresholding or top-ratio filtering to dense primitive-level relevance scores.
However, these scores often vary substantially across scenes and prompts, making such heuristics sensitive and leading to unstable targeting.
ClusterSplat addresses this by reframing 3D target selection as cluster-level selection over scene-adaptive candidates.
The method first learns instance-aware Gaussian features, forms feature-aware seed voxels, and merges neighboring seed voxels when a description-length criterion decreases, producing a scene-specific set of cluster candidates.
Given a referring expression, a query-conditioned cluster scorer ranks the cluster candidates with a lightweight MLP, and the selected cluster directly defines the explicit 3D target while its cluster score is rendered for the 2D mask.
On ScanRefer, ClusterSplat achieves the best rendered 2D segmentation and explicit 3D target metrics among the compared 3DGS baselines. It also obtains the best average Ref-LERF scores, supporting the same trend under fine-grained referring descriptions. On ScanNet v2 class name queries, ClusterSplat achieves the strongest rendered 2D results and the highest 3D matching accuracy, while maintaining competitive 3D IoU.
\pionerevision{Our project page is available at \url{https://jaesung8.github.io/clustersplat/}.}
\end{abstract}

%% file: sections/1_intro.tex
\section{Introduction}

3D Gaussian Splatting (3DGS) has recently emerged as a compelling explicit alternative to neural radiance fields \citep{mildenhall2020nerf}. It offers photorealistic novel-view synthesis with real-time rendering capabilities~\cite{kerbl20233d}. Beyond visual rendering, 3DGS is rapidly evolving into a general-purpose 3D representation capable of structural modeling~\cite{lu2024scaffold, guedon2024sugar} and semantic understanding \citep{qin2024langsplat,wu2024opengaussian,li2025scenesplat,zhu2025objectgs}. A pivotal direction in this evolution is language grounding. This capability enables natural-language querying of 3D scenes through referring expressions. While recent approaches have successfully lifted vision-language embeddings into 3D space for open-vocabulary querying~\cite{kerr2023lerf, qin2024langsplat}, explicitly extracting the precise 3D extent of a target object remains a significant challenge.

The core difficulty lies in the continuous nature of the relevance fields used in prior works. Existing methods typically compute dense, primitive-level relevance scores and isolate the target by applying heuristic discretization, such as global thresholding or top-$k$ filtering~\citep{wu2024opengaussian,drsplat25,qu2024goi}. However, these heuristics are highly sensitive in open-vocabulary settings. Relevance score distributions often vary substantially with the scene context and the complexity of the prompt (e.g., compositional attributes or spatial relations). Consequently, relying on scale-sensitive hyperparameters often leads to unstable targeting. Figure~\ref{fig:teaser} illustrates this failure mode: top-ratio filtering can scatter selected primitives across the referred object and nearby distractors rather than producing one coherent 3D target. As shown by the top-ratio analysis in Appendix~\ref{sec:app_sensitivity}, dense relevance fields often lack clear decision boundaries, causing the target to be missed or merged with background under a fixed selection ratio. Moreover, indoor scenes vary substantially in object count, object scale, clutter, and same-class distractors, so cluster candidates should adapt their granularity to the scene structure. \pionerevision{Section~\ref{sec:cluster_construction_selection} analyzes the candidate-quality trade-off behind this design choice.}

To overcome this instability, we design \textbf{ClusterSplat} as a cluster-level targeting framework (Figure~\ref{fig:overview}) that shifts the selection granularity from individual primitives to scene-adaptive cluster candidates. We view the cluster index as the coarse decision variable for a single referred target and construct these candidates with a scene-adaptive merge procedure guided by description length. These observations, together with the scattered primitive selections in Figure~\ref{fig:teaser}, motivate formulating 3D target selection as a discrete selection problem over coherent object-level candidates rather than as a filtering task over millions of unstructured primitives. The framework can be viewed as two stages that align with the offline scene preparation and query-conditioned grounding blocks in Figure~\ref{fig:overview}. First, we optimize instance-aware Gaussian features and construct feature-aware seed voxels. We then build a candidate graph over seed-voxel descriptors and merge adjacent components only when the description length decreases, producing a scene-specific number of clusters. Second, during target selection, a query-conditioned cluster scorer ranks the candidates with a lightweight MLP to select the target cluster. The selected cluster directly defines the explicit 3D target, and its cluster-level scores are rendered to produce the 2D mask.

Furthermore, we observe that the advancement of 3DGS grounding is hindered by the lack of evaluation protocols that jointly assess rendered 2D masks and explicit 3D target sets. Current benchmarks like Ref-LERF~\cite{he2025refersplat} are limited in scene diversity, while standard semantic segmentation metrics fail to capture the nuances of compositional language grounding. Class name prompts provide category-level grounding, but they cannot uniquely specify one target when multiple objects from the same category appear in a scene. Referring expressions address this ambiguity through attributes and spatial relations, making ScanRefer a more demanding test of object-level targeting. To bridge this gap, we define a reproducible evaluation protocol that combines official ScanNet v2 annotations with ScanRefer expressions~\cite{dai2017scannet,chen2020scanrefer}. Our main comparison uses the subset of 10 scenes adopted by recent 3DGS open-vocabulary baselines for reproducibility. This enables class name grounding and referring expression segmentation to be reported on the same ScanNet v2 scenes.

In summary, our contributions are as follows:
\begin{itemize}
    \item We introduce a scale- and granularity-aware formulation for 3D visual grounding in 3DGS. By transforming primitive-level score filtering into cluster-level selection over scene-adaptive candidates, we avoid per-query score thresholds while allowing the cluster granularity to follow the scene structure.
    \item We propose \textbf{ClusterSplat}, a two-stage framework that first learns instance-aware Gaussian features and constructs scene-adaptive clusters with a description-length criterion, followed by query-conditioned cluster scoring for explicit target selection and 2D mask rendering.
    \item We evaluate ClusterSplat with a reproducible protocol combining ScanNet v2 annotations and ScanRefer expressions that reports class name grounding and referring expression segmentation on the same ScanNet v2 scenes.
    \item On the ScanRefer benchmark, experiments show state-of-the-art referring segmentation among the compared 3DGS baselines. Ref-LERF further confirms consistent gains in fine-grained scenes, while ScanNet v2 class name evaluation shows improved rendered 2D masks and competitive 3D object matching.
\end{itemize}

%% file: sections/2_1_gaussians.tex
\subsection{3D Gaussian Splatting and Language Grounding}

Neural rendering first achieved high-quality novel-view synthesis with implicit radiance fields~\citep{mildenhall2020nerf}, while recent explicit representations such as 3DGS provide real-time rasterization-based rendering~\cite{kerbl20233d}. Explicit primitives also support faster optimization and direct geometric manipulation compared with fully implicit scene encodings. Subsequent work improves 3DGS through structural regularity, filtering, surface-aligned primitives, and alternative optimization strategies~\cite{lu2024scaffold,yu2024mip,huang20242d,kheradmand20243d}.

Beyond rendering, explicit Gaussian primitives have become a substrate for semantic and language-aware feature fields learned through language embeddings, 2D foundation-model distillation, or cross-modal rasterization~\cite{qin2024langsplat,shi2024legaussians,zhou2024feature3dgs,peng2025vlgss,li2025scenesplat,zhu2025objectgs}. However, view-consistent semantic rendering does not by itself define a reliable single-object 3D target. Dr.~Splat directly registers language embeddings to Gaussians~\cite{drsplat25}, interaction-oriented methods lift 2D prompts into Gaussian space~\cite{gaussian_grouping,cen2025segment}, and instance-level Gaussian representations recover object-level subsets~\cite{li2024instancegaussian,piekenbrinck2025opensplat3d}. Yet many pipelines still extract targets from continuous relevance scores using thresholding, top-$k$ selection, or related discretization rules~\citep{qu2024goi,wu2024opengaussian,drsplat25}. These rules remain sensitive to scene- and prompt-dependent score scales, which makes explicit 3D targeting unstable. This sensitivity has also motivated alternatives to fixed similarity cutoffs, such as hyperplane-based selection of Gaussians of Interest~\cite{qu2024goi}.

Referring tasks in 3D Gaussian scenes make this limitation more apparent. ReferSplat introduces the Ref-LERF benchmark and obtains supervision from reliable Grounded-SAM candidates~\citep{he2025refersplat,ren2024grounded}, but view-wise mask cues can still under-specify whether a prediction forms one coherent 3D instance when the Gaussian representation admits ambiguous groupings~\citep{wu2024opengaussian}. Related 3DGS methods use grouped or instance-aware structure for point-level open-vocabulary understanding, view-dependent semantic aggregation, or dense open-vocabulary 3D segmentation~\cite{wu2024opengaussian,cen2025laga,zhu2025cos3d}. These objectives are complementary to our setting, where a referring expression should select one referred target. This motivates grouping primitives into stable candidates and performing discrete selection in the grouped space. In ClusterSplat, clusters serve as the decision domain for single-target referring expressions, and the selected cluster directly determines the explicit 3D target.

%% file: figures/2_overview.tex
\begin{figure}[t]
    \centering
    \includegraphics[width=\linewidth]{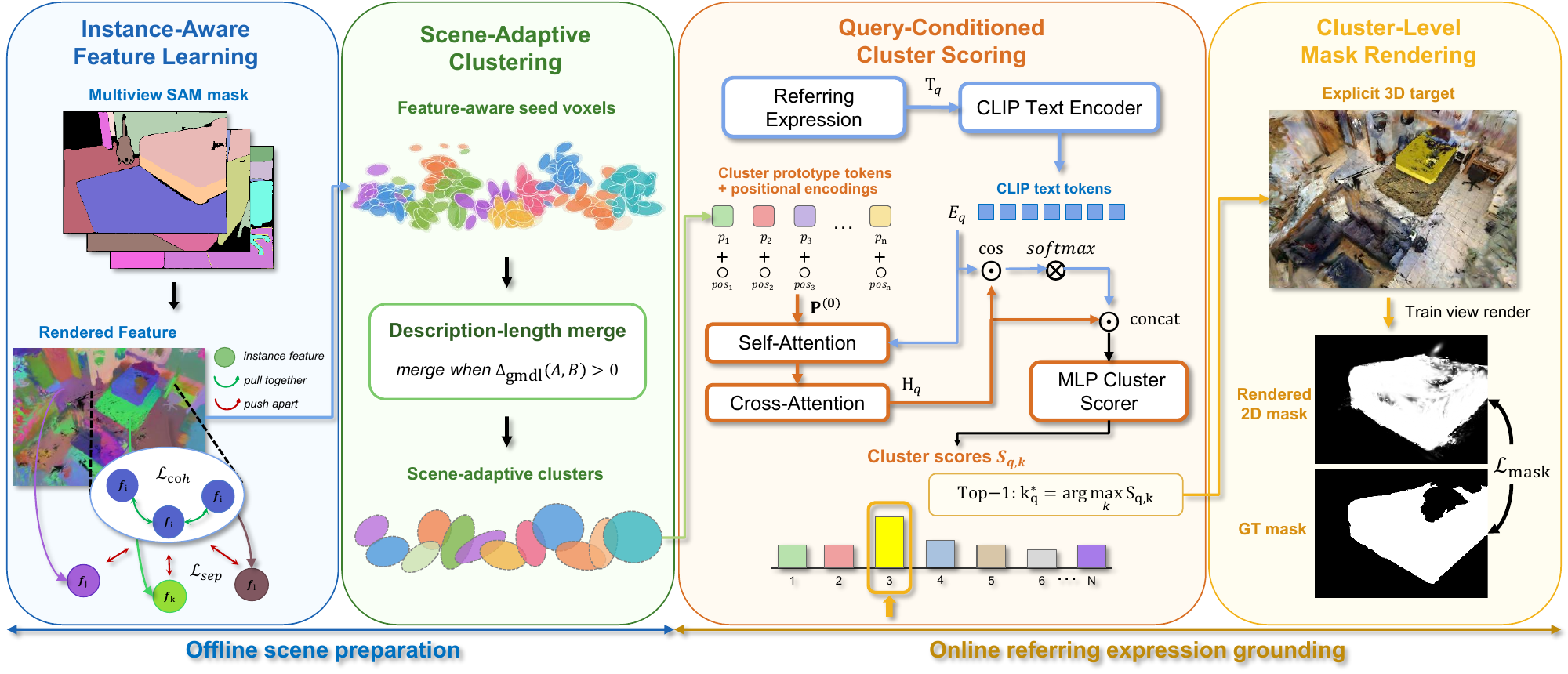}
    \vspace{-5mm}
    \caption{\textbf{ClusterSplat overview.} The pipeline first performs \emph{Instance-Aware Gaussian Feature Learning and Scene-Adaptive Clustering}: it learns instance-aware Gaussian features, forms feature-aware seed voxels, and merges them into scene-adaptive cluster candidates. It then performs \emph{Query-Conditioned Cluster Scoring}: the referring expression conditions the cluster tokens, and the MLP scorer ranks the candidates to select the Top-1 cluster as the explicit 3D target. Finally, \emph{Cluster-Level Mask Rendering} assigns the selected cluster scores to Gaussians and renders a 2D mask supervised by the mask loss.}
    \label{fig:overview}
\end{figure}

%% file: sections/2_2_grounding.tex
\subsection{3D Visual Grounding}

3D visual grounding aims to localize a language-specified target within a 3D scene. It is commonly separated into comprehension tasks, which predict a coarse target location such as a box, and segmentation tasks, which predict a precise point- or mask-level region~\citep{chen2020scanrefer,he2024refmask3d}. ScanNet v2 scenes and ScanRefer expressions have been instrumental in this area by providing RGB-D reconstructions with rich, compositional language descriptions~\citep{dai2017scannet,chen2020scanrefer}. Related 3D referring-expression datasets further study language grounding beyond a single benchmark family~\citep{achlioptas2020referit3d}. Building on these datasets, point-cloud methods have improved language-geometry alignment and relational reasoning~\cite{he2024refmask3d,zhao2021dvgtransformer,wu2023eda}.

Open-vocabulary 3D scene understanding has also been explored with CLIP-aligned point features and class-agnostic 3D mask proposals before the recent shift toward 3DGS representations~\cite{peng_openscene_2023,takmaz2023openmask3d}. Foundation-model supervision further improves scalability, with AugRefer generating diverse synthetic supervision~\cite{wang2025augrefer} and SeeGround enabling zero-shot grounding by bridging 3D scenes with 2D vision-language models~\cite{li2025seeground}.

Adapting this progress to 3DGS requires care because 3DGS optimizes unstructured primitives primarily for rendering. Unlike explicit point clouds, Gaussian primitives are not directly organized as semantic units, so pixel-level parsing can hide failures in the underlying 3D target set. OpenGaussian highlights this 2D-3D discrepancy when improving feature association and discretized 3D selection~\cite{wu2024opengaussian}. Since relevance scores in open-vocabulary 3DGS vary across scenes and prompts, robust evaluation should report both rendered 2D segmentation and discrete 3D target selection for referring expressions. This need motivates our cluster-level formulation, which avoids scale-dependent score discretization while preserving 3D structural consistency and 2D renderability.

%% file: sections/3_1_problem.tex
\subsection{\texorpdfstring{Problem Definition and Target Selection}{Problem Definition and Target Selection}}
\label{sec:problem_overview}

\paragraph{Problem Definition.}
We study 3D visual grounding in a 3DGS scene representation.
A scene is modeled as a set of Gaussian primitives
$\mathcal{G}=\{g_i\}_{i=1}^{N}$.
Each primitive $g_i$ has geometric and appearance parameters
(mean $\mu_i\in\mathbb{R}^{3}$, covariance $\mathbf{\Sigma}_i$, opacity $\alpha_i$, and spherical harmonics $\mathbf{SH}_i$).
We augment each primitive with a learnable instance-aware feature $\mathbf{f}_i\in\mathbb{R}^{D_f}$.
Given a query $q$ with referring expression text $T_q$, the model predicts two outputs.
The first is a 2D segmentation mask for an arbitrary camera pose $\pi$, denoted by $M_{2D}(\pi)$.
The second is a 3D target represented as a subset of Gaussians, $\mathcal{S}_{\mathrm{target}}\subset\mathcal{G}$.
We focus on the single-target referring setting and introduce a query-specific target cluster index $k_q^\star\in\{1,\ldots,K_s\}$.
Here, $K_s$ is the scene-specific number of cluster candidates produced for scene $s$.

\paragraph{3D Targeting via Scene-Adaptive Clusters.}
Prior 3DGS language grounding pipelines typically compute continuous relevance scores over primitives and extract the target using heuristic filtering rules.
However, such primitive-level filtering is inherently unstable as score distributions vary substantially across scenes.
Cluster-level selection provides a more structured decision space, but the cluster candidates must remain faithful to scene-dependent object scale and clutter.
We therefore formulate 3D visual grounding as selection among scene-adaptive cluster candidates.
By aggregating Gaussians into clusters, we reformulate the challenging dense prediction task into a tractable discrete selection problem.
Crucially, this simplification relies on clusters representing coherent object-level candidates rather than arbitrary geometric fragments.
Our cluster candidates are generated by a scene-adaptive merge procedure guided by description length over feature-aware seed voxels defined in Section~\ref{sec:clustering}.
This allows the number of cluster candidates to vary across scenes.
For scene $s$, we partition the Gaussians into $K_s \ll N$ clusters $\{\mathcal{G}_k\}_{k=1}^{K_s}$.
For each query-cluster pair, the scorer in Section~\ref{sec:cluster_attention} produces a query-conditioned score $S_{q,k}$.
The selected index $k_q^\star$ is the cluster index with the maximum score.
The corresponding set $\mathcal{G}_{k_q^\star}$ contains all Gaussians assigned to that selected cluster, and we define the predicted 3D target set as $\mathcal{S}_{\mathrm{target}}=\mathcal{G}_{k_q^\star}$.
The selected cluster directly defines the explicit 3D target, and the cluster-level scores provide the renderable evidence for the 2D mask.

\begin{equation}
k_q^\star=\operatorname*{argmax}_{k\in\{1,\dots,K_s\}} S_{q,k},\qquad
\mathcal{S}_{\mathrm{target}}=\mathcal{G}_{k_q^\star}.
\label{eq:top1_targeting}
\end{equation}

%% file: sections/3_2_clustering.tex
\subsection{\texorpdfstring{Instance-Aware Gaussian Feature Learning and Scene-Adaptive Clustering}{Instance-Aware Gaussian Feature Learning and Scene-Adaptive Clustering}}
\label{sec:clustering}

\paragraph{Instance-aware Gaussian feature learning.}
Stage~1 adapts an instance-aware Gaussian feature objective related to OpenGaussian~\citep{wu2024opengaussian}, using training-view instance masks to learn geometry-aligned Gaussian features.
Each Gaussian has a learnable feature $\mathbf{f}_i\in\mathbb{R}^{D_f}$.
These features are rendered to feature maps with the same opacity-aware compositing used by 3DGS.
Within each training mask, rendered features are pulled toward a mask prototype.
Prototypes from different masks in the same view are separated to discourage feature collapse.
The resulting geometry-aligned feature field is used for scene-adaptive cluster construction.
Appendix~\ref{subsec:mask_feature_details} provides the detailed loss definitions.

\paragraph{Scene-adaptive clustering.}
\pionerevision{After feature learning, we construct cluster candidates from feature-aware seed voxels. Each Gaussian is assigned using its scene-normalized spatial voxel index and the dominant signed dimension of its $\ell_2$-normalized instance feature. This initialization over-segments the scene so Gaussians with different feature directions can remain separate even within the same spatial voxel. For each seed voxel, we compute a normalized feature prototype $\phi_a$ and a 3D centroid. A minimum spanning tree (MST) in the whitened feature-position descriptor space restricts the candidate merges.}

\pionerevision{During greedy merging, a component $S$ is a connected set of seed-voxel prototypes. We evaluate each component with an MDL-inspired description-length cost and a BIC-style component penalty~\citep{rissanen1978modeling,schwarz1978estimating}:}
\begin{equation}
\mathcal{D}_{\mathrm{gmdl}}(S)=
\frac{\mathrm{RSS}_{\mathrm{feat}}(S)}{2\sigma_f^2}
+\lambda_{\mathrm{gmdl}}\frac{1}{2}(D_f+1)\log N_{\mathrm{eff}},
\label{eq:gmdl_cost_short}
\end{equation}
\pionerevision{where $\mathrm{RSS}_{\mathrm{feat}}(S)=|S|-\|\sum_{a\in S}\phi_a\|_2$ measures inconsistency among the feature directions, with each seed prototype contributing unit weight. The feature variance $\sigma_f^2$ and effective prototype count $N_{\mathrm{eff}}$ are computed once per scene. Merging components with inconsistent features increases the first term, while the second term is paid once per component and favors fewer clusters. The coefficient $\lambda_{\mathrm{gmdl}}$ controls this penalty. We merge MST-adjacent components $A$ and $B$ only when the total description length decreases:}
\begingroup
\ifpionereview\color{pionerevisioncolor}\fi
\begin{equation}
\Delta_{\mathrm{gmdl}}(A,B)=
\mathcal{D}_{\mathrm{gmdl}}(A)+\mathcal{D}_{\mathrm{gmdl}}(B)-
\mathcal{D}_{\mathrm{gmdl}}(A\cup B)>0.
\label{eq:gmdl_merge}
\end{equation}
\endgroup
\pionerevision{When no candidate merge reduces the cost, the remaining components define the Gaussian assignment $a(i)\in\{1,\ldots,K_s\}$. The cluster count $K_s$ adapts to each scene, and the assignment remains fixed during Stage~2 training. Appendix~\ref{subsec:adaptive_clustering_details} gives the seed construction, candidate graph, and sufficient-statistic definitions.}

%% file: sections/3_3_scorer.tex
\subsection{\texorpdfstring{Query-Conditioned Cluster Scoring}{Query-Conditioned Cluster Scoring}}
\label{sec:cluster_attention}

From Stage~1 (Section~\ref{sec:clustering}), each Gaussian $g_i$ has an adaptive cluster assignment $a(i)\in\{1,\dots,K_s\}$.
The $k$-th cluster is $\mathcal{G}_k=\{g_i \mid a(i)=k\}$.
Given query $q$ with referring expression text $T_q$, we score the $K_s$ scene clusters and select a 3D target without score-threshold calibration.
We encode $T_q$ with a frozen CLIP text encoder~\citep{radford2021learning}.
The token embeddings are projected to the ClusterSplat hidden dimension.
This yields the token matrix $\mathbf{E}_q=[\mathbf{t}_{q,1},\dots,\mathbf{t}_{q,L_q}]^\top\in\mathbb{R}^{L_q\times d_p}$, where each $\mathbf{t}_{q,\ell}\in\mathbb{R}^{d_p}$.

We summarize cluster geometry by the centroid
\begin{equation}
\bar{\mu}_k=\frac{1}{|\mathcal{G}_k|}\sum_{g_i\in\mathcal{G}_k}\mu_i \in \mathbb{R}^{3}.
\label{eq:cluster_centroid}
\end{equation}

We associate each cluster with a learnable cluster token $\mathbf{c}_k\in\mathbb{R}^{d_p}$.
We inject 3D position using sinusoidal encoding $\mathrm{PE}(\cdot)$ \cite{vaswani2017attention}:
\begin{equation}
\begin{aligned}
\mathbf{p}_k^{(0)} &= \mathbf{c}_k+\mathrm{PE}(\bar{\mu}_k),\\
\mathbf{P}^{(0)} &= [\mathbf{p}_1^{(0)},\dots,\mathbf{p}_{K_s}^{(0)}]^\top \in \mathbb{R}^{K_s\times d_p}.
\end{aligned}
\label{eq:init_proto}
\end{equation}

We update the initialized cluster tokens $\mathbf{P}^{(0)}$ using a Transformer-decoder architecture~\citep{vaswani2017attention,carion2020end}.
First, we apply self-attention to model the global spatial and semantic context:
\begin{equation}
    \mathbf{P}^{(1)} = \text{LN}(\mathbf{P}^{(0)} + \text{SA}(\mathbf{P}^{(0)})).
\end{equation}
A 3D target is often distinguished by its relationship with surrounding distractors.
Self-attention therefore contextualizes each cluster token with the scene layout.
This context reduces false-positive clusters that share local appearance cues but lack the correct relational support.

Subsequently, we employ cross-attention to align the contextualized cluster tokens with CLIP token features $\mathbf{E}_q$:
\begin{equation}
    \mathbf{H}_q = \text{LN}(\mathbf{P}^{(1)} + \text{CA}(\mathbf{P}^{(1)}, \mathbf{E}_q)).
\end{equation}
Here, $\mathbf{H}_q$ is the query-conditioned token matrix for all scene candidates, with one row per cluster and hidden dimension $d_p$.
Unlike simple global pooling, this cross-attention step allows each cluster representation to access the query tokens that are relevant to that candidate.
These tokens may encode object nouns, attributes, or spatial prepositions.

\paragraph{MLP cluster scorer.}
Let $\mathbf{h}_{q,k}$ denote the query-conditioned token for the $k$-th cluster candidate, corresponding to the $k$-th row of $\mathbf{H}_q$.
To preserve token-level query information, the scorer forms a cluster-specific query summary with cosine-similarity token attention,
\begin{equation}
\beta_{q,k,\ell}=\mathrm{softmax}_{\ell}\!\left(
\frac{\mathrm{cos}(\mathbf{h}_{q,k},\mathbf{t}_{q,\ell})}{\tau}
\right),\qquad
\mathbf{y}_{q,k}=\sum_{\ell=1}^{L_q}\beta_{q,k,\ell}\mathbf{t}_{q,\ell}.
\label{eq:token_query_summary}
\end{equation}
Here, $\mathrm{cos}(\mathbf{a},\mathbf{b})=\langle\mathbf{a}/\|\mathbf{a}\|_2,\mathbf{b}/\|\mathbf{b}\|_2\rangle$, and $\tau$ is the token-attention temperature.
\pionerevision{A two-layer MLP with a 128-dimensional hidden layer predicts the score from the cluster token, its query summary, and their elementwise product ($\odot$) and absolute difference:}
\begingroup
\ifpionereview\color{pionerevisioncolor}\fi
\begin{equation}
S_{q,k}=\mathrm{MLP}_{\mathrm{score}}\!\left(
[\mathbf{h}_{q,k};\mathbf{y}_{q,k};\mathbf{h}_{q,k}\odot\mathbf{y}_{q,k};|\mathbf{h}_{q,k}-\mathbf{y}_{q,k}|]
\right).
\label{eq:mlp_cluster_score}
\end{equation}
\endgroup
This scorer preserves the final decision as a relative ranking over cluster candidates rather than a threshold on primitive-level scores.
We use these scores as the cluster-level evidence for the query-specific target cluster index introduced in Section~\ref{sec:problem_overview}.
We select the top-scoring cluster and define the explicit 3D target set:
\begin{equation}
\begin{aligned}
k_q^\star &= \operatorname*{argmax}_{k\in\{1,\dots,K_s\}} S_{q,k},\\
\mathcal{S}_{\mathrm{target}} &= \{g_i\in\mathcal{G}\mid a(i)=k_q^\star\}.
\end{aligned}
\label{eq:top1_select}
\end{equation}

This argmax is the decision rule of the single-target cluster formulation.
It relies on the relative ranking of structured cluster candidates.
This avoids per-query or per-scene thresholds on primitive-level score magnitudes.

%% file: sections/3_4_rendering.tex
\subsection{Cluster-Level Mask Rendering}
\label{sec:mask_rendering}

After query-conditioned scoring, each Gaussian inherits the score of its assigned cluster,
\begin{equation}
    \ell_{q,i}=S_{q,a(i)}.
\label{eq:cluster_gaussian_logit}
\end{equation}
The explicit 3D output is the Top-1 cluster set in Eq.~\eqref{eq:top1_select}.
For the 2D output at camera pose $\pi$, we render the Gaussian logits $\ell_{q,i}$ by the same opacity-aware Gaussian rasterization used for 3DGS rendering. For pixel $u$,

\begin{equation}
R_q(\pi,u)
=
\sum_{i=1}^{N}
\ell_{q,i}\,
\alpha_{\pi,i}(u)\prod_{j=1}^{i-1}\bigl(1-\alpha_{\pi,j}(u)\bigr),
\qquad
P_q(\pi,u)=\sigma(R_q(\pi,u)).
\label{eq:cluster_mask_render}
\end{equation}

Given the training-view ground-truth mask $M_q(\pi)$, Stage~2 is supervised by the same focal and Dice objective used in our implementation,

\begin{equation}
\mathcal{L}_{\mathrm{mask}}
=
(1-\lambda_D)\mathcal{L}_{\mathrm{focal}}\!\big(P_q(\pi),M_q(\pi)\big)
+\lambda_D\mathcal{L}_{\mathrm{dice}}\!\big(P_q(\pi),M_q(\pi)\big),
\qquad \lambda_D=0.5.
\label{eq:cluster_mask_loss}
\end{equation}

Thus, cluster scores are learned through differentiable 2D mask supervision while the explicit 3D prediction remains the selected Gaussian cluster~\cite{lin2017focal,milletari2016v}.

%% file: tables/1_benchmarks.tex
\begin{table}[t]
    \centering
    \caption{\pionerevision{ScanRefer and ScanNet v2 quantitative results on the shared evaluation scenes. All scores are percentages. Bold and underlined values indicate the best and second-best distinct scores in each column.}}
    \label{table:main_benchmarks}
    \setlength{\tabcolsep}{4.2pt}

    \resizebox{\linewidth}{!}{
    \begin{tabular}{l cc cc cc cc}
        \toprule
        \multirow{3}{*}{Methods}
            & \multicolumn{4}{c}{ScanRefer}
            & \multicolumn{4}{c}{ScanNet v2 class name} \\
        \cmidrule(lr){2-5}\cmidrule(lr){6-9}
            & \multicolumn{2}{c}{2D}
            & \multicolumn{2}{c}{3D}
            & \multicolumn{2}{c}{2D}
            & \multicolumn{2}{c}{3D} \\
        \cmidrule(lr){2-3}\cmidrule(lr){4-5}\cmidrule(lr){6-7}\cmidrule(lr){8-9}
            & mIoU$\uparrow$ & mAcc@0.25$\uparrow$
            & mIoU$\uparrow$ & mAcc@0.25$\uparrow$
            & mIoU$\uparrow$ & mAcc@0.25$\uparrow$
            & mIoU$\uparrow$ & mAcc$\uparrow$ \\
        \midrule
        LangSplat~\cite{qin2024langsplat}
            & 11.00 & 13.41 & 5.81 & 4.95
            & 27.67 & 40.82 & 2.01 & 8.13 \\
        OpenGaussian~\cite{wu2024opengaussian}
            & 18.48 & 5.53 & 1.06 & 0.10
            & 19.55 & 26.11 & \textbf{31.05} & \underline{46.39} \\
        Dr.Splat~\cite{drsplat25}
            & 19.01 & 27.76 & \underline{9.48} & \underline{11.17}
            & 19.94 & 32.64 & \underline{30.67} & 46.12 \\
        ReferSplat~\cite{he2025refersplat}
            & 31.30 & 45.07 & 4.05 & 0.10
            & 28.70 & 40.12 & 6.35 & 17.22 \\
        ReferSplat (w/ our loss)
            & \underline{31.55} & \underline{45.99} & 3.85 & 0.00
            & \underline{32.70} & \underline{47.32} & 4.52 & 12.89 \\
        ClusterSplat (Ours)
            & \pionerevision{\textbf{39.72}} & \pionerevision{\textbf{54.35}} & \pionerevision{\textbf{23.33}} & \pionerevision{\textbf{43.38}}
            & \textbf{42.20} & \textbf{58.30} & 29.21 & \textbf{46.60} \\
        \bottomrule
    \end{tabular}
    }

\end{table}

%% file: figures/3_scanrefer.tex
\begin{figure}[t]
    \centering
    \includegraphics[width=\linewidth]{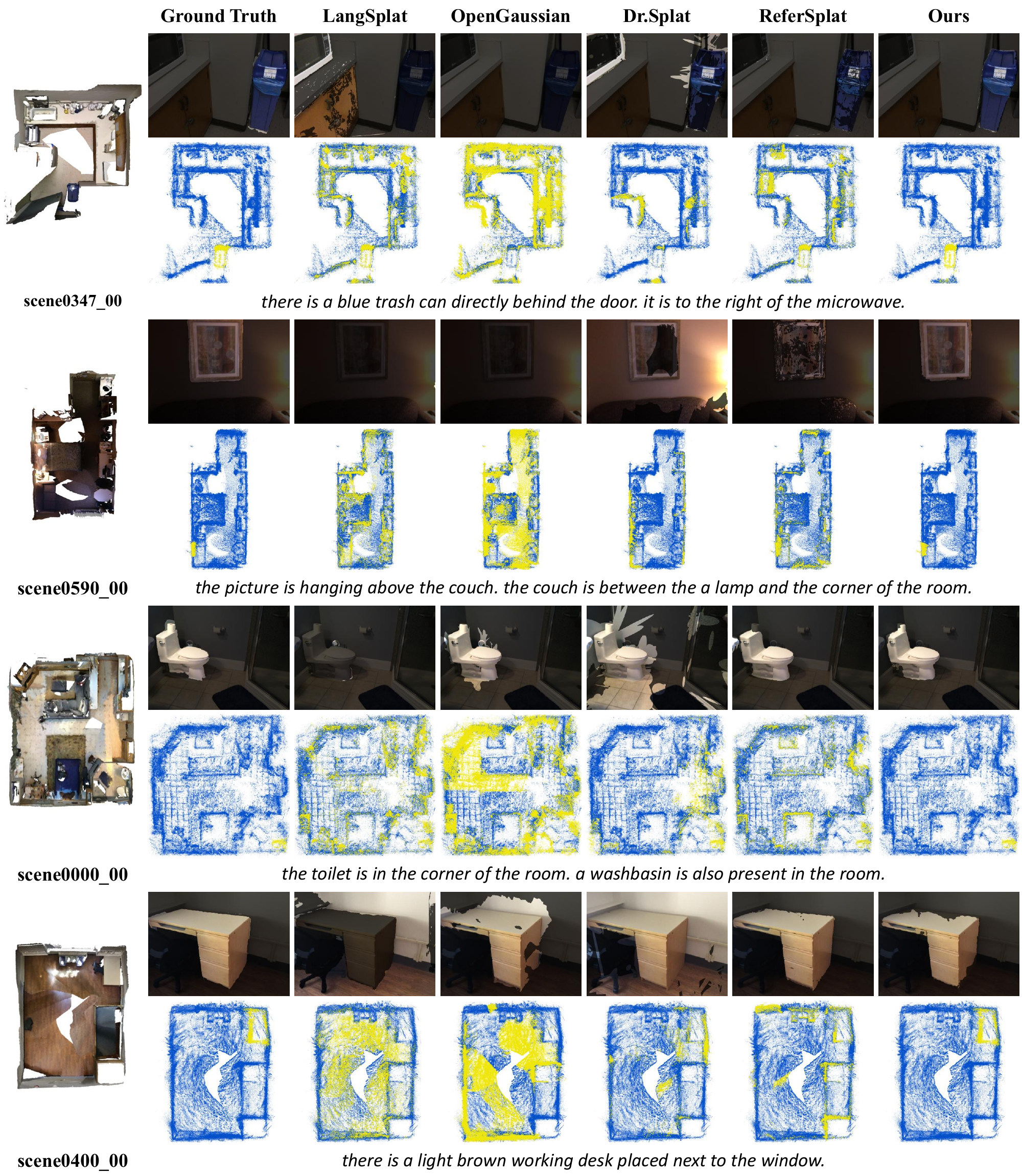}
    \vspace{-5mm}
    \caption{Qualitative examples on ScanRefer. Each row shows a referring expression with rendered masks and top-down 3D selections. ClusterSplat localizes the referred object more coherently and reduces distractor support compared with score-discretized baselines.}
    \label{fig:scanrefer}
\end{figure}

%% file: sections/4_1_setup.tex
\subsection{Experimental Setup}
\label{sec:exp_setup}

\paragraph{Datasets.}
To evaluate 3D visual grounding in 3DGS, we use two complementary benchmarks. We first pair ScanNet v2~\cite{dai2017scannet} RGB-D scenes and official annotations with ScanRefer~\cite{chen2020scanrefer} referring expressions, using the scene subset and preprocessing convention adopted by recent 3DGS open-vocabulary evaluations. ScanRefer provides referring expressions and target object IDs for objects in the ScanNet v2 scenes, enabling evaluation of both class name grounding and referring expression segmentation within the same reconstructed scenes. We also use Ref-LERF from ReferSplat~\cite{he2025refersplat} to assess fine-grained spatial reasoning in small-scale scenes; Appendix~\ref{subsec:data_prep} gives the ReferSplat confidence-weighted IoU pseudo-mask rule used for Ref-LERF supervision.

\paragraph{Evaluation Protocol.}
For ScanNet v2 and ScanRefer, we evaluate on the common 10-scene subset adopted by OpenGaussian~\cite{wu2024opengaussian}. Table~\ref{table:main_benchmarks} reports two complementary settings on the same reconstructed scenes: ScanRefer referring segmentation and ScanNet v2 class name grounding. Both settings evaluate rendered 2D masks and explicit 3D Gaussian target sets. Appendix~\ref{subsec:data_prep} and Appendix~\ref{sec:appendix_supervision} provide the held-out view split, annotation sources, and object ID propagation details.

\paragraph{Baselines.}
We compare ClusterSplat against language-embedded 3DGS pipelines, including LangSplat~\citep{qin2024langsplat}, OpenGaussian~\citep{wu2024opengaussian}, Dr.Splat~\citep{drsplat25}, and ReferSplat~\citep{he2025refersplat}. Prior works typically output continuous primitive relevance scores, so they require thresholding or top-ratio discretization to define a 3D target set. We therefore sweep the top-ratio in $\{0.01,0.02,\ldots,0.10\}$ for each score-based baseline and report the best per-method oracle result; ClusterSplat instead uses Top-1 cluster selection and does not tune a per-query or per-scene score threshold.

%% file: sections/4_2_results.tex
\subsection{\pionerevision{Benchmark results}}
\label{sec:main_results}

\paragraph{ScanRefer referring segmentation.}
ScanRefer directly evaluates the referred-object setting because each referring expression is tied to a specific object ID in a ScanNet v2 scene. Table~\ref{table:main_benchmarks} therefore reports ScanRefer first. Compared with ScanNet v2 class name evaluation, ScanRefer is more demanding because the model must identify one referred object among category distractors using attributes, spatial relations, and instance-specific language. Under this setting, ClusterSplat improves both rendered 2D segmentation and explicit 3D target selection over the compared 3DGS baselines. The comparison with ReferSplat trained under the same supervision indicates that the gain is not solely attributable to pseudo-caption or mask supervision, but also to the cluster-level selection architecture. The larger improvement in 3D target metrics further suggests that this architecture is especially useful when the query contains attributes or spatial relations that make primitive-level score filtering unstable.

\paragraph{ScanNet v2 class name evaluation.}
The ScanNet v2 class name results in Table~\ref{table:main_benchmarks} provide a complementary diagnostic on the same evaluation scenes. ClusterSplat substantially improves rendered 2D segmentation, indicating that the selected clusters provide stable image-space evidence across held-out views. We include this evaluation because prior language-embedded 3DGS studies commonly report 3D evaluation based on class name queries, and the results show that ClusterSplat remains competitive under this established protocol. In 3D, ClusterSplat attains the highest matching accuracy and remains close to the strongest 3D IoU baselines. Since class name prompts describe categories rather than a unique referred instance, this protocol has limited ability to test targeting of a specific object in scenes with multiple objects from the same category. This limitation motivates the ScanRefer referring segmentation evaluation above.

\paragraph{Ref-LERF.}
\pionerevision{On Ref-LERF, ClusterSplat reaches 31.14\% average mIoU and 41.70\% mAcc@0.25, compared with 26.59\% and 36.35\% for ReferSplat with our loss. Using the same human expressions and pseudo-masks, this gain supports cluster-level selection under matched supervision and loss. Appendix~\ref{subsec:reflerf_results} reports the per-scene results in Table~\ref{table:ref-lerf}.}

\paragraph{Qualitative analysis.}
Figure~\ref{fig:scanrefer} supports the ScanRefer trends with queries that require object identity and relational cues, such as a trash can behind a door, a picture above a couch, a toilet in a room corner, and a desk next to a window. Score-discretized baselines often leak support to nearby structures or distractor regions in the rendered view and the top-down 3D selection. ClusterSplat keeps the selected support concentrated on the referred object and yields masks aligned with the target.

\paragraph{\pionerevision{Efficiency.}}
\pionerevision{On an RTX A6000, ClusterSplat takes 19.55\,ms per query after 125.3\,min of method-specific scene preparation. Preparation is incurred once per scene, whereas queries are repeated during interaction. Appendix~\ref{sec:efficiency_analysis} reports the comparison and timing scope.}

%% file: sections/4_3_analysis.tex
\subsection{\pionerevision{Design analysis}}
\label{sec:cluster_construction_selection}

\pionerevision{Table~\ref{tab:cluster_quality_gmdl} shows why purity alone is insufficient for single-cluster grounding. Increasing $K$ in fixed KMeans raises purity but lowers Coverage@1 and Oracle Top-1 IoU as objects split across more clusters. Even perfect selection then recovers less of the target. Scene-adaptive clustering offers more complete candidates, achieving the highest Coverage@1 and Oracle Top-1 IoU despite lower purity than fixed KMeans. This tradeoff matters because Stage~2 can rank candidates but cannot repair their boundaries. Appendix~\ref{subsec:cluster_quality} defines the metrics.}

\input{tables/2_clusters.tex}

\pionerevision{Table~\ref{tab:cluster_cumulative_topk} tests whether multi-cluster selection can compensate for this fragmentation. We retrain Stage~2 under the same protocol for each construction and form cumulative Top-$k$ unions from each model. Fixed KMeans with $K=256$ and $K=512$ benefits from multiple selections, consistent with fragment recovery. Scene-adaptive clustering instead falls from 23.33\% 3D mIoU at Top-1 to 14.96\% at Top-5, suggesting that the cost of false positives outweighs the benefit of fragment recovery. This agrees with the focal and Dice supervision in Section~\ref{sec:mask_rendering}, which encourages query-associated foreground responses and penalizes distractors through rendered cluster logits. No tested union exceeds scene-adaptive Top-1, supporting this combination for the evaluated single-target queries.}

\input{tables/3_topk.tex}

\paragraph{\pionerevision{Text encoder and caption source.}}
\pionerevision{Appendix~\ref{subsec:ablation} compares frozen CLIP and BERT~\citep{devlin2019bert} encoders under the same ClusterSplat configuration. CLIP achieves 23.33\% 3D mIoU, compared with 14.62\% for BERT. This supports CLIP for our cluster scorer, consistent with a benefit from its image-text pretraining. The caption-source comparison in Appendix~\ref{subsec:caption_sensitivity} gives 23.33\% for GPT-4.1, 21.23\% for InternVL3.5-8B~\citep{wang2025internvl35}, and 20.01\% for Qwen3-VL-8B~\citep{bai2025qwen3vl}. Both local caption generators outperform the compared ScanRefer baselines. The gains persist across the tested generators, although their magnitude depends on the caption source.}

%% file: tables/2_clusters.tex
\begin{table}[!htbp]
\centering
\caption{\pionerevision{Cluster candidate quality before Stage~2 on ScanRefer.} Purity, Coverage@1, and Oracle Top-1 IoU are fractions in $[0,1]$; Frag@0.9 is a cluster count. \pionerevision{Metric definitions are in Appendix~\ref{subsec:cluster_quality}.} Bold values indicate the best score in each column.}
\label{tab:cluster_quality_gmdl}
\setlength{\tabcolsep}{4.0pt}
\renewcommand{\arraystretch}{1.08}

\resizebox{\linewidth}{!}{%
\begin{tabular}{l c c c c}
\toprule
Method & Purity $\uparrow$ & Coverage@1 $\uparrow$ & \shortstack{Oracle Top-1\\IoU $\uparrow$} & Frag@0.9 $\downarrow$ \\
\midrule
Fixed KMeans ($K=128$) & 0.7825 & 0.5649 & 0.4017 & 5.95 \\
Fixed KMeans ($K=256$) & 0.8094 & 0.4840 & 0.3734 & 8.90 \\
Fixed KMeans ($K=512$) & \textbf{0.8269} & 0.3935 & 0.3200 & 14.15 \\
Fixed KMeans + pair merge & 0.7436 & 0.6693 & 0.4615 & 4.71 \\
\textbf{Scene-adaptive clustering (ours)} & 0.7635 & \textbf{0.6933} & \textbf{0.4798} & 3.86 \\
Scene-adaptive clustering + xyz cost & 0.7745 & 0.6720 & 0.4682 & \textbf{3.85} \\
\bottomrule
\end{tabular}}

\end{table}

%% file: tables/3_topk.tex
\begin{table}[!htbp]
\centering
\caption{\pionerevision{Effect of cluster construction and cumulative Top-$k$ selection on ScanRefer 3D mIoU (\%). Within each construction, Top-$k$ predictions combine the $k$ highest-scoring clusters from the same trained model. Bold indicates the highest value in the table.}}
\label{tab:cluster_cumulative_topk}
\setlength{\tabcolsep}{5pt}
\renewcommand{\arraystretch}{1.08}
\begin{tabular}{l c c c c c}
\toprule
\pionerevision{Construction} & \pionerevision{Top-1} & \pionerevision{Top-2} & \pionerevision{Top-3} & \pionerevision{Top-4} & \pionerevision{Top-5} \\
\midrule
\pionerevision{Fixed KMeans ($K=128$)} & \pionerevision{18.31} & \pionerevision{17.66} & \pionerevision{16.42} & \pionerevision{15.81} & \pionerevision{14.47} \\
\pionerevision{Fixed KMeans ($K=256$)} & \pionerevision{15.44} & \pionerevision{16.34} & \pionerevision{16.27} & \pionerevision{16.32} & \pionerevision{16.21} \\
\pionerevision{Fixed KMeans ($K=512$)} & \pionerevision{10.97} & \pionerevision{13.29} & \pionerevision{14.56} & \pionerevision{14.60} & \pionerevision{14.62} \\
\pionerevision{Fixed KMeans + pair merge} & \pionerevision{21.23} & \pionerevision{19.66} & \pionerevision{18.67} & \pionerevision{17.47} & \pionerevision{16.46} \\
\pionerevision{Scene-adaptive clustering} & \pionerevision{\textbf{23.33}} & \pionerevision{19.78} & \pionerevision{17.42} & \pionerevision{16.09} & \pionerevision{14.96} \\
\pionerevision{Scene-adaptive + xyz cost} & \pionerevision{21.63} & \pionerevision{19.64} & \pionerevision{17.41} & \pionerevision{16.09} & \pionerevision{14.56} \\
\bottomrule
\end{tabular}
\end{table}

%% file: sections/5_discussion.tex
\section{Discussion and Limitations}
\label{sec:discussion}

ClusterSplat reduces dependence on primitive-level score filtering by selecting a scene-adaptive cluster as the explicit target. Its main limitation is therefore the candidate pool. Over-merging nearby similar objects or splitting one object across several clusters can limit the conservative Top-1 selection. This explains why ScanNet v2 class name matching accuracy is strong while exact 3D IoU remains stricter: the selected cluster can overlap the correct object enough for matching, yet still miss parts of the object extent or include nearby fragments. Future work can handle fragmented candidates with uncertainty-aware multi-cluster selection.

\pionerevision{Our formulation handles expressions that refer to a single object. Separate queries can select several objects, but grounding a plural expression that refers to a variable number of instances remains future work.}

\pionerevision{The experiments cover ten indoor ScanNet v2 scenes and four Ref-LERF scenes, with each scene prepared separately. Broader scene coverage and transfer to unseen scenes remain to be evaluated. Training relies on annotated instance masks and generated descriptions, and each new scene requires offline preparation before queries can be processed.}

%% file: sections/6_conclusion.tex
\section{Conclusion}
\label{sec:conclusion}

We presented ClusterSplat, which converts 3DGS visual grounding from primitive-level score filtering into discrete selection over scene-adaptive Gaussian clusters. The method learns instance-aware Gaussian features, builds scene-adaptive clusters with a description-length criterion, ranks them with query-conditioned scoring, and renders cluster-level scores for 2D mask prediction. Results show state-of-the-art referring segmentation on ScanRefer among the compared 3DGS baselines, with Ref-LERF providing complementary fine-grained validation. The method also produces strong rendered masks and competitive 3D object matching on ScanNet v2. These results support scene-adaptive cluster selection for referring expression grounding in 3DGS. Remaining errors suggest future work on uncertainty-aware multi-cluster selection while preserving explicit target sets.

%% file: sections/7_ack.tex
\begingroup
\ifpionereview
\renewenvironment{ack}{\acksection}{}
\color{pionerevisioncolor}
\fi
\begin{ack}
This work was supported by the National Research Foundation of Korea (NRF) grant funded by the Korea government (MSIT) (RS-2025-00573160); the Technology Innovation Program (RS-2025-02222776, Development and Demonstration of AI and Lightweight Technology-Based Automated E-Waste Sorting and Retrieval System) funded by the Ministry of Trade, Industry and Resources (MOTIR, Korea); and the ``Advanced GPU Utilization Support Program'' funded by the Government of the Republic of Korea (Ministry of Science and ICT).
\end{ack}
\endgroup

%% file: appendix/0_intro.tex
\section*{Supplementary Material}

In this supplementary material, we provide additional details omitted from the manuscript.
Section~\ref{sec:implementation} covers implementation details and the evaluated 3D tasks. Section~\ref{sec:additional_results} includes analysis of baseline methods, ablation studies, and qualitative results.

%% file: appendix/A_1_setup.tex
\subsection{System Setup and Hyperparameters}
\label{subsec:system_setup}
\label{sec:appendix_setup}
Unless otherwise specified, experiments are conducted on a single NVIDIA RTX 5090 GPU. The online efficiency measurements in \pionerevision{Appendix~\ref{sec:efficiency_analysis}} are measured on an NVIDIA RTX A6000 server. We implement our framework using PyTorch 2.8.0+cu128 and CUDA 12.8. To ensure training stability and precision, mixed-precision training is not employed.

For the model architecture, we use a frozen CLIP text encoder~\citep{radford2021learning} and project token embeddings to the ClusterSplat cluster/text hidden dimension $d_p=64$. Cluster assignments are loaded from the offline scene-adaptive feature-clustering step and used as fixed assignments during Stage~2. The resulting scene-specific cluster counts on the 10-scene ScanNet v2 subset are 53-143, with an average of 87.3 clusters per scene.

The token-conditioned cluster scorer is a two-layer MLP with one hidden layer of dimension 128, separate from the prototype dimension $d_p$. The same cluster scores are used for Top-1 3D target selection and rendered 2D mask prediction.

The model is optimized using the Adam optimizer with a fixed learning rate of $1 \times 10^{-4}$ for all experiments. \pionerevision{The total number of training iterations is set to 10,000.} We use the focal and Dice segmentation objective with equal weights and do not use cluster-entropy regularization in the current full setting.

%% file: appendix/A_2_features.tex
\subsection{\texorpdfstring{Instance-Aware Gaussian Feature Learning Details}{Instance-Aware Gaussian Feature Learning Details}}
\label{subsec:mask_feature_details}

Stage~1 learns instance-aware Gaussian features from mask supervision to build the feature field used by scene-adaptive clustering.
For this feature-learning branch, we follow the OpenGaussian-style setup and obtain mask-level supervision from multi-view SAM instance masks~\citep{kirillov2023segment,wu2024opengaussian}, separate from the ScanNet v2 masks used for pseudo-caption and evaluation-label construction.
For a training view $v$ and pixel $u\in\Omega$, the per-Gaussian feature $\mathbf{f}_i$ is rendered with opacity-aware compositing:
\begin{equation}
\mathbf{F}_v(u)
=
\sum_{i=1}^{N}
\mathbf{f}_i\,
\alpha_{v,i}(u)\prod_{j=1}^{i-1}\bigl(1-\alpha_{v,j}(u)\bigr),
\label{eq:render_feat_unorm}
\end{equation}
where $\alpha_{v,i}(u)$ is the per-pixel opacity contribution of Gaussian $i$ after front-to-back sorting.
We use the normalized rendered feature $\tilde{\mathbf{F}}_v(u)=\mathbf{F}_v(u)/\|\mathbf{F}_v(u)\|_2$ for the mask-level losses.

Let $\mathcal{M}_v=\{M_{v,m}\}_{m=1}^{M_v}$ denote training-view instance masks and $\Omega_{v,m}=\{u\in\Omega\mid M_{v,m}(u)=1\}$ denote the pixel support of mask $m$.
The normalized prototype of mask $m$ is
\begin{equation}
\begin{aligned}
\bar{\boldsymbol{\psi}}_{v,m} &= \frac{1}{|\Omega_{v,m}|}\sum_{u\in\Omega_{v,m}} \tilde{\mathbf{F}}_v(u),\\
\boldsymbol{\psi}_{v,m} &= \frac{\bar{\boldsymbol{\psi}}_{v,m}}{\|\bar{\boldsymbol{\psi}}_{v,m}\|_2}.
\end{aligned}
\label{eq:mask_prototype}
\end{equation}
Following OpenGaussian-style mask supervision~\citep{wu2024opengaussian}, the objective combines intra-mask cohesion and inter-mask separation:
\begin{equation}
\mathcal{L}_{\mathrm{coh}}
=
\frac{1}{\sum_v M_v}\sum_v\sum_{m=1}^{M_v}
\frac{1}{|\Omega_{v,m}|}
\sum_{u\in\Omega_{v,m}}
\left\|\tilde{\mathbf{F}}_v(u)-\boldsymbol{\psi}_{v,m}\right\|_2^2,
\label{eq:mask_cohesion}
\end{equation}
\begin{equation}
\mathcal{L}_{\mathrm{sep}}
=
\frac{1}{\sum_v M_v(M_v-1)}
\sum_v\sum_{m\ne n}
\frac{w_{v,mn}}{\|\boldsymbol{\psi}_{v,m}-\boldsymbol{\psi}_{v,n}\|_2^2+\epsilon},
\label{eq:mask_separation}
\end{equation}
where $w_{v,mn}$ is a distance-rank weight for nearby or easily confused mask pairs.
The Stage~1 instance-feature loss is
\begin{equation}
\mathcal{L}_{\mathrm{inst}}
=\lambda_{\mathrm{coh}}\mathcal{L}_{\mathrm{coh}}
+\lambda_{\mathrm{sep}}\mathcal{L}_{\mathrm{sep}}.
\label{eq:instance_feature_loss}
\end{equation}

%% file: appendix/A_3_clustering.tex
\subsection{\texorpdfstring{Scene-Adaptive Clustering and Mask Rendering Details}{Scene-Adaptive Clustering and Mask Rendering Details}}
\label{subsec:adaptive_clustering_details}

This subsection expands the mathematical details of the scene-adaptive clustering stage used by the reported model. The clustering stage is performed once per scene before query-conditioned training. It uses Gaussian centers and learned instance features, but not the evaluation query, CLIP text features, 3D object IDs, or a manually fixed target number of clusters.

\paragraph{Feature-aware seed voxels.}
Let $\tilde{\mu}_i\in[0,1]^3$ denote the per-scene normalized center of Gaussian $g_i$, and let $\hat{\mathbf{f}}_i=\mathbf{f}_i/\|\mathbf{f}_i\|_2$ denote its normalized instance feature. Given voxel size $\delta$, the spatial voxel index is
\begin{equation}
    \nu_i=\left\lfloor \tilde{\mu}_i/\delta \right\rfloor .
\end{equation}
We further split each occupied voxel by the dominant signed feature direction,
\begin{equation}
    d_i=\operatorname*{argmax}_{j} |\hat{\mathbf{f}}_{i,j}|,
    \qquad
    s_i=\mathrm{sign}(\hat{\mathbf{f}}_{i,d_i}),
    \qquad
    z_i=(\nu_i,d_i,s_i).
\end{equation}
The key $z_i$ defines the initial seed voxel assignment. This construction intentionally over-segments the scene so that small objects and local boundaries are less likely to be absorbed before merging.

\paragraph{Seed-voxel prototypes and candidate graph.}
For each seed voxel $a$, we compute a normalized feature prototype and a spatial centroid,
\begin{equation}
    \phi_a=\frac{\sum_{i:z_i=a}\hat{\mathbf{f}}_i}
    {\left\|\sum_{i:z_i=a}\hat{\mathbf{f}}_i\right\|_2},
    \qquad
    \bar{\mu}_a=\frac{1}{n_a}\sum_{i:z_i=a}\tilde{\mu}_i,
\end{equation}
where $n_a=|\{i:z_i=a\}|$. In the final setting, each seed voxel contributes one prototype observation with unit weight. Candidate merge edges are then built from a whitened feature-position descriptor,
\begin{equation}
    \mathbf{y}_a=[\phi_a;\bar{\mu}_a],
    \qquad
    \tilde{\mathbf{y}}_a=(\mathbf{y}_a-\boldsymbol{m}_y)\oslash\boldsymbol{s}_y,
    \qquad
    E_{\mathrm{cand}}=\mathrm{MST}(\{\tilde{\mathbf{y}}_a\}_a),
\end{equation}
where $\boldsymbol{m}_y$ and $\boldsymbol{s}_y$ are the per-dimension mean and standard deviation, and $\oslash$ denotes element-wise division. This graph restricts which neighboring prototypes can be tested for merging, while the description-length cost below decides whether an allowed pair should merge.

\paragraph{Description-length merge.}
We adopt an MDL-inspired description-length objective for components built from seed-voxel prototypes; its complexity term adapts a BIC-style asymptotic penalty to this component representation~\citep{rissanen1978modeling,schwarz1978estimating}.
For a component $S$ containing seed-voxel prototypes, let
\begin{equation}
    W_S=\sum_{a\in S}w_a,
    \qquad
    \mathbf{G}_S=\sum_{a\in S}w_a\phi_a,
    \qquad
    \mathrm{RSS}_{\mathrm{feat}}(S)=W_S-\|\mathbf{G}_S\|_2.
\end{equation}
With unit prototype weights in the reported model, $w_a=1$. The scene-global feature variance and effective count are
\begin{equation}
    \sigma_f^2=\max\!\left(
    \frac{W_{\mathrm{all}}-\|\mathbf{G}_{\mathrm{all}}\|_2}
    {W_{\mathrm{all}}D_f},\epsilon\right),
    \qquad
    N_{\mathrm{eff}}=\max\!\left(
    \frac{W_{\mathrm{all}}^2}{\sum_a w_a^2},1\right).
\end{equation}
The component cost used for merging is
\begin{equation}
    \mathcal{D}_{\mathrm{gmdl}}(S)=
    \frac{\mathrm{RSS}_{\mathrm{feat}}(S)}{2\sigma_f^2}
    +\lambda_{\mathrm{gmdl}}\frac{1}{2}(D_f+1)\log N_{\mathrm{eff}},
\end{equation}
up to constants that cancel in pairwise merge decisions. The second term has the standard $\frac{p}{2}\log N$ information-criterion form with $p=D_f+1$ and $N=N_{\mathrm{eff}}$, but we use it as an MDL-inspired penalty rather than an exact BIC score. The second term is paid once per component, so a candidate edge $(A,B)$ is accepted only when
\begin{equation}
    \Delta_{\mathrm{gmdl}}(A,B)=
    \mathcal{D}_{\mathrm{gmdl}}(A)+\mathcal{D}_{\mathrm{gmdl}}(B)
    -\mathcal{D}_{\mathrm{gmdl}}(A\cup B)>0.
\end{equation}
Accepted merges add the sufficient statistics of the two components and update the candidate graph. The final labels are compacted into an assignment $a(i)\in\{1,\ldots,K_s\}$, where $K_s$ is determined by the accepted merges and can vary across scenes.

\paragraph{Cluster scores to mask supervision.}
During Stage~2, the assignment is fixed. Given cluster scores $S_{q,k}$, Gaussian $g_i$ receives logit $\ell_{q,i}=S_{q,a(i)}$. For camera pose $\pi$, opacity-aware rendering produces a logit map
\begin{equation}
    R_q(\pi)=\mathrm{Render}_{\pi}\bigl(\{\ell_{q,i}\}_{i=1}^{N}\bigr),
    \qquad
    P_q(\pi)=\sigma(R_q(\pi)).
\end{equation}
The Stage~2 mask objective is
\begin{equation}
    \mathcal{L}_{\mathrm{mask}}
    =(1-\lambda_D)\mathcal{L}_{\mathrm{focal}}(P_q,M_q)
    +\lambda_D\mathcal{L}_{\mathrm{dice}}(P_q,M_q),\qquad \lambda_D=0.5,
\end{equation}
where $M_q$ is the training-view target mask. Thus, the query-conditioned scorer is trained through rendered 2D mask supervision while the explicit 3D output remains the selected cluster.

%% file: appendix/A_4_5_data.tex
\subsection{Data Preparation}
\label{subsec:data_prep}
\textbf{ScanNet v2 Data Formatting.} We preprocess the ScanNet v2 sequences by decoding the raw \texttt{.sens} files to extract RGB images and camera poses using the official ScanNet toolkit. For fair comparison and reproducibility, we use the 10-scene evaluation subset introduced by OpenGaussian and follow the standard data processing pipeline for 3DGS-based methods.

\textbf{Interleaved Train/Evaluation Split.} To rigorously evaluate 3D generalization on unseen viewpoints, we adopt an interleaved evaluation split. We define the evaluation set $\mathcal{V}_{eval}$ as every 20th frame with an offset of 10. The remaining frames constitute the training set $\mathcal{V}_{train}$. All pseudo-caption generation and feature supervision construction are performed after this split and only on $\mathcal{V}_{train}$, so the held-out evaluation frames are never used to construct training supervision.

\textbf{Evaluation Protocol Details.} The main quantitative table for ScanNet v2 and ScanRefer reports two related but distinct evaluations on the same reconstructed scenes. The ScanRefer columns use referring expressions tied to target object instances, while the ScanNet v2 columns use class name queries derived from official scene annotations. For 2D evaluation, predicted masks are rendered on held-out frames and compared with official ScanNet v2 instance annotations projected to those frames~\citep{dai2017scannet}. For 3D evaluation, object instance IDs are propagated to Gaussians using the Mahalanobis-distance assignment described below, and IoU is computed over the resulting object ID Gaussian set~\cite{drsplat25}. ScanRefer target object IDs define the foreground instance for referring segmentation queries, preventing same-category distractors from being counted as target foreground.

\textbf{Ground Truth Generation.} We employ distinct supervision strategies for training and evaluation:
\begin{itemize}
    \item Training set ($\mathcal{V}_{train}$): In ScanNet v2 and ScanRefer experiments, we use the official ScanNet v2 instance masks on training frames as visual anchors and generate pseudo-captions for those masks using the \textit{Automated Supervision Generation Pipeline} detailed in Sec.~\ref{sec:appendix_supervision}.
    \item Evaluation set ($\mathcal{V}_{eval}$): To strictly evaluate rendered 2D segmentation quality, we utilize the official ScanNet v2 2D instance annotation projections (\texttt{*\_2d-instance-filt.zip}). For 3D referring segmentation, we use ScanNet v2 and ScanRefer object instance IDs to construct object-level Gaussian pseudo-GT labels, as detailed in Sec.~\ref{sec:appendix_supervision}.
\end{itemize}

\textbf{Ref-LERF pseudo masks.}
For Ref-LERF, we follow ReferSplat's Grounded-SAM pseudo-mask selection protocol~\citep{he2025refersplat,ren2024grounded}.
Given $K$ candidate masks $\{\hat{M}_k\}_{k=1}^{K}$ and confidences $\{\gamma_k\}_{k=1}^{K}$ for an image-expression pair, define the confidence-weighted mask $A_k(u)=\gamma_k\hat{M}_k(u)$.
The ReferSplat confidence-weighted IoU score is
\begin{equation}
 p_k=\sum_{j=1}^{K}
 \frac{\sum_{u}\min\big(A_k(u),A_j(u)\big)}
 {\sum_{u}\max\big(A_k(u),A_j(u)\big)},
 \qquad
 \hat{M}=\hat{M}_{\operatorname*{argmax}_{k}p_k}.
\label{eq:reflerf_confidence_weighted_iou}
\end{equation}
Equivalently, this is the weighted-overlap form $\sum_j (\gamma_k\hat{M}_k\cap\gamma_j\hat{M}_j)/(\gamma_k\hat{M}_k\cup\gamma_j\hat{M}_j)$ proposed in ReferSplat.
\subsection{Automated Pseudo Label Pipeline}
\label{subsec:auto_supervision}
\label{sec:appendix_supervision}
We generate language supervision for the training views ($\mathcal{V}_{train}$) using training frame ScanNet v2 instance masks and Vision-Language Models (VLMs). The masks are official training frame annotations, while the associated captions are pseudo-captions generated for language supervision. This setting is therefore annotation-assisted rather than annotation-free. Evaluation frame masks, official ScanRefer referring expressions, and evaluation target object IDs are not provided to the pseudo-caption generator or the optimization objective.

\textbf{Step 1: Training-frame Instance Mask Extraction.} For each training frame, we extract the official ScanNet v2 2D instance masks and compute the tight 2D bounding box for each visible instance. These training frame masks, rather than SAM proposals, serve as the visual anchors for the subsequent caption generation process.

\textbf{Step 2: Extensive Pseudo-Caption Generation.} Previous works have explored the use of VLMs to generate comprehensive 3D object and scene descriptions at scale \cite{wang2025augrefer}. Similarly, we employ GPT-4.1 \cite{gpt41} to generate five diverse pseudo-captions for each training frame instance mask. The GPT-4.1 input consists of the training RGB image, the instance mask, and its bounding box; it does not include official ScanRefer expressions, evaluation target object IDs, evaluation-frame masks, or 3D object annotations. Unlike 3D visual grounding methods that rely on a single description per object, these multiple pseudo-captions provide richer linguistic variation while remaining independent of the evaluation queries. We prompt GPT-4.1 to describe visual attributes, spatial context, and object distinctiveness visible within the masked region, following the types of attributes emphasized in 3D visual grounding \cite{li2025seeground}. As shown by the generated captions in Figure \ref{fig:pseudo_caption_generation}, this produces pseudo-captions that describe the training frame instance without exposing the held-out target answer.

\input{figures/4_captions.tex}

\textbf{Leakage-safe supervision set.} Using ScanNet v2 instance masks on training frames is leakage-safe with respect to our held-out evaluation protocol as long as the split is applied before supervision is constructed. Specifically, for a training view $v\in\mathcal{V}_{train}$, let $\mathcal{M}^{inst}_v=\{M^{inst}_{v,m}\}_{m=1}^{M_v}$ be the official instance masks visible in that training frame and let $c^{(r)}_{v,m}$ denote the $r$-th GPT pseudo-caption for mask $m$. The training caption-supervision set is
\[
\mathcal{D}^{cap}_{train}=
\{(v,M^{inst}_{v,m},c^{(r)}_{v,m}):v\in\mathcal{V}_{train}, m\in[1,M_v], r\in[1,R]\}.
\]
For a ScanRefer evaluation query $q$ with expression $T_q$ and target object ID $o_q$, the official expression $T_q$ and held-out evaluation masks are not included in $\mathcal{D}^{cap}_{train}$. The object ID $o_q$ may correspond to the same physical instance visible in training frames, which is expected in a scene-level view split protocol; however, it is not supplied to GPT-4.1 as the query answer or referring expression during caption generation. The 2D evaluation mask is constructed only on held-out frames from the official instance projection $A^{2D}_v$ as $M^{2D}_{q,v}(u)=\mathbf{1}[A^{2D}_v(u)=o_q]$, for $v\in\mathcal{V}_{eval}$. Thus, the protocol is not annotation-free, but it avoids leakage from evaluation views and official referring expressions into training.

\textbf{Object ID Gaussian pseudo-GT for 3D evaluation.} For 3D referring segmentation, the target should be the referred object instance rather than every object of the same semantic class. We therefore follow the Gaussian label propagation protocol based on Mahalanobis distance used in Dr.~Splat~\cite{drsplat25}, but replace the semantic label space with ScanNet v2 and ScanRefer object IDs. Let $\mathcal{P}_s=\{p_k\}_{k=1}^{Q}$ be annotated points in scene $s$, each with object ID $o^{p_k}\in\mathcal{O}_s$, and let $g_i$ be a Gaussian with center $\mu_i$ and covariance $\mathbf{\Sigma}_i$. We compute
\[
d^{\mathrm{mahal}}(p_k,g_i)=(p_k-\mu_i)^\top\mathbf{\Sigma}_i^{-1}(p_k-\mu_i),
\qquad
\eta(p_k,g_i)=\exp(-\tfrac{1}{2}d^{\mathrm{mahal}}(p_k,g_i)).
\]
The object ID pseudo-GT label of Gaussian $g_i$ is assigned by aggregating this local affinity over annotated points with the same object ID,
\[
o_i^g=\operatorname*{argmax}_{o\in\mathcal{O}_s}
\sum_{p_k\in\mathcal{P}_s}\mathbf{1}[o^{p_k}=o]\,\eta(p_k,g_i).
\]
Given a ScanRefer query with expression $T_q$ and target ID $o_q$, the 3D pseudo-GT foreground is $\mathcal{Y}^{\mathrm{obj}}_q=\{i:o_i^g=o_q\}$. This object ID definition prevents same-class distractors from being counted as foreground and is therefore aligned with referring expression segmentation. For qualitative inspection, Figure~\ref{fig:sup_scannet_19class} visualizes the resulting Gaussian pseudo labels alongside the annotated ScanNet v2 point cloud labels after mapping object IDs to the 19 semantic classes.

\textbf{Template-based Augmentation for ScanNet v2.} While GPT-generated captions provide linguistic diversity, short category names in ScanNet v2 (e.g., "chair", "table") can be too terse as language prompts. To provide the CLIP text encoder with natural sentence context, we wrap the class nouns into simple prompt templates:
\begin{itemize}
    \item "There is a \{\} in the scene."
    \item "The scene contains a \{\}."
    \item "This visual element is part of a \{\}."
    \item "This local feature corresponds to a \{\}."
\end{itemize}
This augmentation preserves class name supervision in a sentence-like form and improves the stability of text-cluster alignment without introducing evaluation queries into training.

%% file: figures/4_captions.tex
\begin{figure}[t!]
    \centering
    \includegraphics[width=\columnwidth]{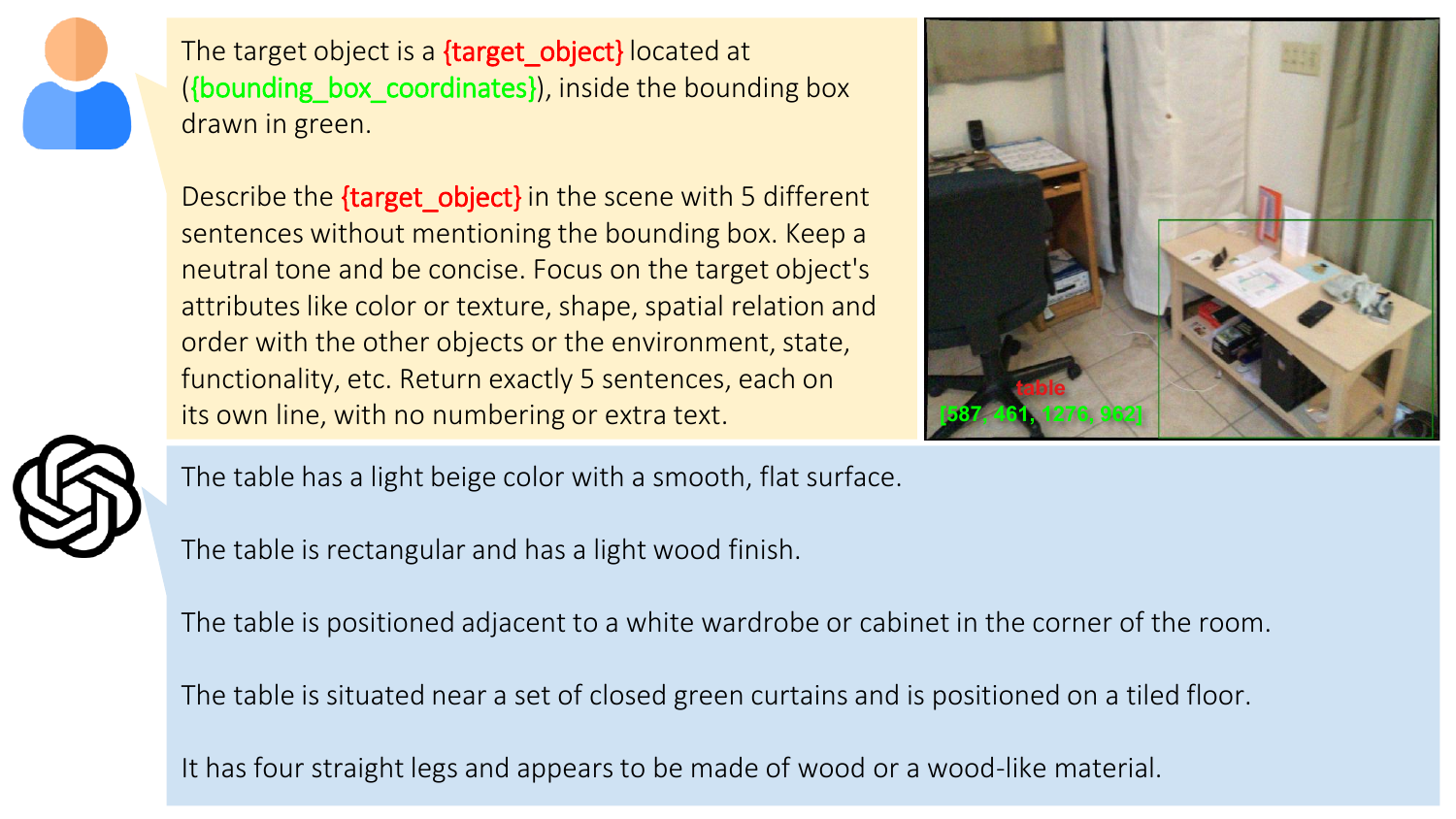}
    \vspace{-5mm}
    \caption{
      Illustration of how we prompt GPT-4.1 to obtain five diverse pseudo-captions for a training-frame instance mask. The specific view shows two tables, and the mask/bounding-box cue enables GPT-4.1 to generate captions that are descriptive and localized to the selected instance.
    }
    \label{fig:pseudo_caption_generation}
\end{figure}

%% file: appendix/A_6_baselines.tex
\subsection{Baseline Implementation Details}
\label{subsec:baseline_impl}
\label{sec:appendix_baselines}
We evaluate our framework against four strong language-embedded 3DGS pipelines: LangSplat \cite{qin2024langsplat}, OpenGaussian \cite{wu2024opengaussian}, Dr.Splat \cite{drsplat25}, and ReferSplat \cite{he2025refersplat}. To ensure a fair comparison, all baselines are reproduced using their official implementations with the following specific configurations:
\begin{itemize}
    \item \textbf{LangSplat:} We employ the CLIP-based auto-encoder with a latent dimension of 3.
    \item \textbf{OpenGaussian:} We use the official implementation with the SAM-based feature setting and evaluate its continuous relevance scores under the same held-out view split and top-ratio sweep described below.
    \item \textbf{Dr.Splat:} We use the official implementation with the released configuration on the same reconstructed scenes. Its continuous relevance scores are converted to 3D target sets using the same top-ratio sweep, while the Mahalanobis label propagation protocol is used only to construct object ID pseudo-GT for evaluation.
    \item \textbf{ReferSplat:} We train this model using the same training-frame mask and pseudo-caption supervision as our method. This ensures that any performance gains are attributed to our architectural contributions rather than differences in training data quality.
\end{itemize}

\textbf{3D Target Extraction Strategy.}
A critical challenge in benchmarking is that prior works typically output continuous relevance scores, requiring a discretization step to define the 3D target set $\mathcal{S}_{\mathrm{target}}$. As shown in Fig.~\ref{fig:top_ratio}, a fixed global threshold is often suboptimal due to scale variations across scenes.
To prevent underestimating the baselines, we adopt an oracle evaluation protocol. Specifically, for each baseline method, we perform a grid search for the \textit{top-ratio} parameter $\rho\in\{0.01,0.02,\ldots,0.10\}$. We report the best per-method oracle performance achieved across this range. This strategy provides a per-method oracle within the swept range, so the comparison is less affected by under-tuned discretization choices for competing score-based methods.

%% file: appendix/B_0_results.tex
\section{Additional Results}
\label{sec:additional_results}

This section provides further analysis of the experiments. \pionerevision{We examine selection heuristics, model design, cluster candidates, caption sources, and lexical overlap, then give per-scene Ref-LERF results and additional qualitative comparisons.}

%% file: appendix/B_1_sensitivity.tex
\subsection{Sensitivity Analysis of Selection Heuristics}
\label{subsec:sensitivity}
\label{sec:app_sensitivity}

We investigated the stability of primitive-level score discretization in existing baselines. To this end, we conducted a grid search on the \textbf{top-ratio} hyperparameter for OpenGaussian~\cite{wu2024opengaussian} and Dr.Splat~\cite{drsplat25}. We evaluated the 3D targeting performance across 10 ScanNet v2 scenes by sweeping the top-ratio from $0.01$ to $0.10$.

Figure~\ref{fig:top_ratio} visualizes the results. We observe that the optimal selection ratio depends heavily on the scene. For example, Dr.Splat peaks at a ratio of $0.04$ in \texttt{scene0070}. However, \texttt{scene0645} shows a decreasing trend. This variation indicates that using a fixed global ratio is ineffective for open-vocabulary scenarios. Additionally, the performance curves show sharp peaks. Small deviations from the optimal ratio cause significant drops in 3D IoU. These findings confirm that primitive-level relevance fields lack clear decision boundaries in 3D space. Our method uses a Top-1 cluster extraction rule that removes per-query and per-scene score-threshold calibration during 3D target extraction.

\input{figures/5_sensitivity.tex}

%% file: figures/5_sensitivity.tex
\begin{figure}[ht]
    \centering
    \includegraphics[width=1.0\textwidth]{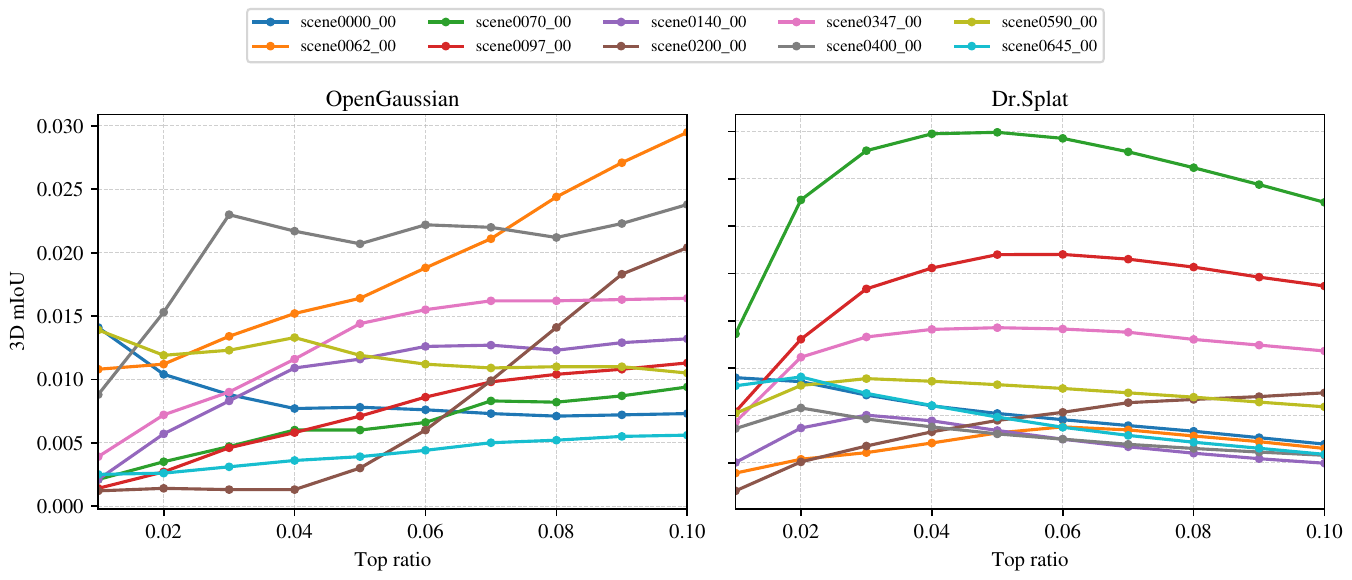}
    \caption{\textbf{Sensitivity of primitive-level target extraction to the top-ratio hyperparameter.}
    We evaluate the 3D targeting performance (mIoU) of OpenGaussian~\citep{wu2024opengaussian} and Dr.Splat~\citep{drsplat25}
    across 10 ScanNet v2 scenes by sweeping the selection ratio $\rho \in [0.01, 0.10]$.
    The optimal ratio is highly scene-dependent, and the performance often drops sharply away from the peak,
    indicating that primitive-level score discretization requires careful calibration for robust open-vocabulary 3D targeting.
    This motivates cluster-level Top-1 (argmax) selection, which relies on relative cluster ranking rather than absolute score thresholds.}
    \label{fig:top_ratio}
\end{figure}

%% file: appendix/B_2_design.tex
\subsection{Design Choices}
\label{subsec:ablation}
\label{sec:app_ablation_param}

\pionerevision{We examine the text encoder, cluster candidates, scorer modules, and prototype dimension.}

\paragraph{\pionerevision{Comparison settings.}}
\label{subsec:control_protocol}
\pionerevision{The encoder, caption-source, and cluster-construction comparisons use $d_p=64$ on the ten ScanRefer scenes. Cumulative Top-$k$ predictions combine score-ranked clusters from the same trained model.}

\paragraph{Text backbone scope.}
ClusterSplat uses CLIP token features at the cluster level. Adaptive cluster tokens attend to query tokens and are ranked by a lightweight MLP scorer.
\pionerevision{Table~\ref{tab:clip_bert_control} compares frozen CLIP ViT-B/32 and BERT-large-uncased with the same scene features, cluster assignments, scorer, loss, and training settings. Both encoders provide last-layer token features projected to $d_p=64$ and use their respective tokenization. CLIP yields higher scores across all reported metrics. Its image-text pretraining~\citep{radford2021learning} may make its tokens better suited to alignment with visual cluster representations through projection, cross-attention, and cosine token attention. The paired comparison supports CLIP for this scorer; the preferred encoder should be evaluated within each grounding architecture.}

\input{tables/4_encoder.tex}

\paragraph{Adaptive cluster candidates.}
The cluster assignment is generated before Stage~2 and then kept fixed as cluster labels during query-conditioned training. This design keeps the online target selection rule simple: the model ranks a scene-specific set of renderable cluster candidates and selects the Top-1 cluster. \pionerevision{Table~\ref{tab:cluster_quality_gmdl} evaluates this cluster candidate pool before query-conditioned training; Appendix~\ref{subsec:cluster_quality} defines the metrics.}

\paragraph{Cluster-level rendering.}
The cluster scorer ranks scene-adaptive candidates, the Top-1 cluster defines the explicit 3D target, and the corresponding cluster-level scores are rendered for 2D mask prediction. This keeps the inference rule aligned with the central cluster-level formulation.

\paragraph{Module ablation.}
Table~\ref{tab:scanrefer_module_ablation} reports a ScanRefer ablation of the CLIP-based cluster scorer while keeping the prototype dimension fixed at $d_p=64$ for all variants. The token scorer baseline already gives a strong 3D mIoU, so the gains are not simply due to adding more layers. Cross-attention alone does not improve the 3D metrics, but the full scorer with position encoding, cluster self-attention, cross-attention, and token scoring gives the best 2D accuracy, 3D mIoU, 3D mAcc@0.25, and Hit@1. Overall, the components are most useful as a joint scorer for scene-level cluster ranking rather than as independent monotonic additions. We analyze the effect of $d_p$ separately in Table~\ref{tab:prototype_dim_ablation}.

\input{tables/5_modules.tex}

\paragraph{Prototype dimension.}
Table~\ref{tab:prototype_dim_ablation} varies the prototype dimension $d_p$ while keeping the full scorer. The 128-dimensional setting gives the highest rendered 2D scores, but its explicit 3D target metrics are lower than those of $d_p=64$. Increasing the dimension to 256 further reduces 3D mIoU, mAcc@0.25, mAcc, and Hit@1. We therefore use $d_p=64$, which provides the best 3D target selection while retaining competitive 2D segmentation.

\input{tables/6_dimension.tex}

%% file: tables/4_encoder.tex
\begin{table}[H]
\centering
\caption{\pionerevision{Frozen-encoder comparison on ScanRefer. Bold values indicate the best score in each column.}}
\label{tab:clip_bert_control}
\setlength{\tabcolsep}{6pt}
\renewcommand{\arraystretch}{1.08}
\begin{tabular}{l c c c c}
\toprule
\multirow{2}{*}{\pionerevision{Encoder}} & \multicolumn{2}{c}{\pionerevision{2D}} & \multicolumn{2}{c}{\pionerevision{3D}} \\
\cmidrule(lr){2-3}\cmidrule(lr){4-5}
 & \pionerevision{mIoU$\uparrow$} & \pionerevision{mAcc@0.25$\uparrow$} & \pionerevision{mIoU$\uparrow$} & \pionerevision{mAcc@0.25$\uparrow$} \\
\midrule
\pionerevision{BERT} & \pionerevision{31.47} & \pionerevision{43.48} & \pionerevision{14.62} & \pionerevision{25.40} \\
\pionerevision{CLIP} & \pionerevision{\textbf{39.72}} & \pionerevision{\textbf{54.35}} & \pionerevision{\textbf{23.33}} & \pionerevision{\textbf{43.38}} \\
\bottomrule
\end{tabular}
\end{table}

%% file: tables/5_modules.tex
\begin{table}[H]
    \centering
    \caption{Module ablation on ScanRefer with $d_p=64$ for all variants. Bold values indicate the best score in each column.}
    \label{tab:scanrefer_module_ablation}
    \setlength{\tabcolsep}{4.0pt}
    
    \resizebox{\linewidth}{!}{
    \begin{tabular}{l c c c c cc ccc}
        \toprule
        \multirow{2}{*}{Variant}
            & \multirow{2}{*}{Pos.}
            & \multirow{2}{*}{Cluster SA}
            & \multirow{2}{*}{Cross-attn}
            & \multirow{2}{*}{Token scorer}
            & \multicolumn{2}{c}{2D}
            & \multicolumn{3}{c}{3D} \\
        \cmidrule(lr){6-7}\cmidrule(lr){8-10}
            & & & &
            & mIoU$\uparrow$ & mAcc@0.25$\uparrow$
            & mIoU$\uparrow$ & mAcc@0.25$\uparrow$ & Hit@1$\uparrow$ \\
        \midrule
        Pooled scorer baseline
            & - & - & - & -
            & 37.93 & 51.59 & 22.20 & 41.29 & 43.52 \\
        Token scorer baseline
            & - & - & - & \checkmark
            & 37.86 & 51.66 & 23.17 & 41.96 & 42.29 \\
        + Cross-attention
            & - & - & \checkmark & \checkmark
            & 37.77 & 52.18 & 21.69 & 40.75 & 42.67 \\
        + Cluster self-attention
            & - & \checkmark & \checkmark & \checkmark
            & 37.81 & 51.69 & 22.55 & 41.34 & 44.05 \\
        Full scorer
            & \checkmark & \checkmark & \checkmark & \checkmark
            & \textbf{39.72} & \textbf{54.35} & \textbf{23.33} & \textbf{43.38} & \textbf{46.03} \\
        \bottomrule
    \end{tabular}
    }
    
\end{table}

%% file: tables/6_dimension.tex
\begin{table}[H]
    \centering
    \caption{Prototype dimension ablation on ScanRefer with the full scorer. Bold values indicate the best score in each column.}
    \label{tab:prototype_dim_ablation}
    \setlength{\tabcolsep}{5.0pt}
    
    \resizebox{0.82\linewidth}{!}{
    \begin{tabular}{c cc cccc}
        \toprule
        \multirow{2}{*}{$d_p$}
            & \multicolumn{2}{c}{2D}
            & \multicolumn{4}{c}{3D} \\
        \cmidrule(lr){2-3}\cmidrule(lr){4-7}
            & mIoU$\uparrow$ & mAcc@0.25$\uparrow$
            & mIoU$\uparrow$ & mAcc@0.25$\uparrow$ & mAcc$\uparrow$ & Hit@1$\uparrow$ \\
        \midrule
        32  & 38.75 & 52.59 & 22.06 & 40.92 & 33.66 & 43.88 \\
        64  & 39.72 & 54.35 & \textbf{23.33} & \textbf{43.38} & \textbf{35.86} & \textbf{46.03} \\
        128 & \textbf{40.41} & \textbf{55.00} & 21.42 & 40.32 & 32.09 & 41.71 \\
        256 & 39.77 & 54.78 & 20.52 & 36.69 & 30.30 & 39.42 \\
        \bottomrule
    \end{tabular}
    }
    
\end{table}

%% file: appendix/B_3_clusters.tex
\subsection{\pionerevision{Cluster Evaluation Protocols}}
\label{subsec:cluster_quality}
\label{sec:app_cluster_quality}

We further analyze the cluster candidates before query-conditioned Stage~2 training. This analysis uses object ID labels assigned to 3D Gaussians and therefore measures the quality of the cluster candidates themselves, not the final query-conditioned segmentation performance. We report four complementary metrics. Purity measures whether a cluster is dominated by one object, Coverage@1 measures how much of a ground-truth object is covered by its best single cluster, Oracle Top-1 IoU measures the best possible single-cluster IoU under perfect cluster selection, and Frag@0.9 measures how many clusters are needed to cover 90\% of an object.

\pionerevision{Section~\ref{sec:cluster_construction_selection} relates these candidate metrics in Table~\ref{tab:cluster_quality_gmdl} to downstream grounding in Table~\ref{tab:cluster_cumulative_topk}.}

\paragraph{\pionerevision{Cumulative selection protocol.}}
\pionerevision{We also evaluate downstream grounding under the training settings in Appendix~\ref{subsec:control_protocol}. Let $k_{q,1},\ldots,k_{q,K_s}$ denote the cluster indices ranked by query scores. The cumulative Top-$k$ prediction is}
\[
\ifpionereview\color{pionerevisioncolor}\fi
\mathcal{S}^{(k)}_q=\bigcup_{j=1}^{k}\mathcal{G}_{k_{q,j}}.
\]
\pionerevision{Thus, Top-3 evaluates the union of three ranked clusters, not the third cluster alone. Table~\ref{tab:cluster_cumulative_topk} reports mean ScanRefer 3D mIoU over the ten scenes. The same trained model is used for all five unions within each construction.}

%% file: appendix/B_4_captions.tex
\subsection{\pionerevision{Caption Source and Evaluation-Language Analysis}}
\label{subsec:caption_sensitivity}

\paragraph{\pionerevision{Caption-generator sensitivity.}}
\pionerevision{We compare captions from GPT-4.1, InternVL3.5-8B, and Qwen3-VL-8B using the same scene features, cluster candidates, CLIP encoder, and training settings described in Appendix~\ref{subsec:control_protocol}. Table~\ref{tab:caption_generator_control} shows that GPT-4.1 gives the highest scores. The lower scores with InternVL3.5-8B and Qwen3-VL-8B show sensitivity to the caption source. Both local models still outperform the compared ScanRefer baselines in Table~\ref{table:main_benchmarks}, so the gains persist with local caption generation.}

\input{tables/7_captions.tex}

\paragraph{\pionerevision{Scene-local lexical comparison.}}
\pionerevision{We compare 20,290 training captions with 958 official ScanRefer evaluation expressions from the ten scenes. Each expression is compared with all training captions from the same scene. After Unicode-NFKC normalization, lowercasing, and punctuation removal, we compute token-set Jaccard similarity:}
\[
\ifpionereview\color{pionerevisioncolor}\fi
J(A_q,A_c)=\frac{|A_q\cap A_c|}{|A_q\cup A_c|},\qquad
J_q^{\mathrm{nearest}}=\max_{\text{same-scene }c}J(A_q,A_c),
\]
\pionerevision{where $A_q$ and $A_c$ are the distinct token sets of an expression and a caption. No stopword removal or stemming is applied for this metric. The median nearest-caption Jaccard similarity is 0.391.}

\pionerevision{In a separate analysis, we remove stopwords and stem alphabetic content words before comparing each expression with the training-caption vocabulary of its scene. Of the 958 evaluation expressions, 286 (29.85\%) contain a content word absent from that vocabulary. The evaluation set thus includes expressions with wording and content words beyond the scene's training captions. These lexical statistics do not measure grounding accuracy on that subset separately.}

%% file: tables/7_captions.tex
\begin{table}[H]
\centering
\caption{\pionerevision{Caption-source comparison on ScanRefer. Bold values indicate the best score in each column.}}
\label{tab:caption_generator_control}
\setlength{\tabcolsep}{6pt}
\renewcommand{\arraystretch}{1.08}
\begin{tabular}{l c c c c}
\toprule
\multirow{2}{*}{\pionerevision{Caption generator}} & \multicolumn{2}{c}{\pionerevision{2D}} & \multicolumn{2}{c}{\pionerevision{3D}} \\
\cmidrule(lr){2-3}\cmidrule(lr){4-5}
 & \pionerevision{mIoU$\uparrow$} & \pionerevision{mAcc@0.25$\uparrow$} & \pionerevision{mIoU$\uparrow$} & \pionerevision{mAcc@0.25$\uparrow$} \\
\midrule
\pionerevision{GPT-4.1} & \pionerevision{\textbf{39.72}} & \pionerevision{\textbf{54.35}} & \pionerevision{\textbf{23.33}} & \pionerevision{\textbf{43.38}} \\
\pionerevision{InternVL3.5-8B} & \pionerevision{38.60} & \pionerevision{52.82} & \pionerevision{21.23} & \pionerevision{39.00} \\
\pionerevision{Qwen3-VL-8B} & \pionerevision{34.22} & \pionerevision{46.63} & \pionerevision{20.01} & \pionerevision{36.82} \\
\bottomrule
\end{tabular}
\end{table}

%% file: appendix/B_5_reflerf.tex
\subsection{\pionerevision{Ref-LERF Quantitative Results}}
\label{subsec:reflerf_results}

\pionerevision{Table~\ref{table:ref-lerf} gives the per-scene results summarized in Section~\ref{sec:main_results}. Ref-LERF complements the room-scale ScanRefer evaluation with four small-scale scenes and fine-grained referring expressions. ClusterSplat achieves the highest average mIoU and mAcc@0.25. The gains on scenes such as \textit{teatime} suggest that selecting a coherent cluster provides more stable object support under detailed descriptions.}

\input{tables/8_reflerf.tex}

%% file: tables/8_reflerf.tex
\begin{table}[H]
    \centering
    \caption{Ref-LERF quantitative results in percentages. Bold and underlined values indicate the best and second-best distinct scores in each column; ties receive the same marking. ClusterSplat improves the average referring segmentation scores across the evaluated scenes.}
    \label{table:ref-lerf}
    \resizebox{1.0\linewidth}{!}{
    \renewcommand{\arraystretch}{1.15}
    \begin{tabular}{l|ccccc|ccccc}
        \toprule
        \multirow{2}{*}{Methods}
            & \multicolumn{5}{c|}{mIoU$\uparrow$}
            & \multicolumn{5}{c}{mAcc@0.25$\uparrow$} \\
        & ramen & figurines & teatime & kitchen & Avg.
        & ramen & figurines & teatime & kitchen & Avg. \\
        \midrule
        LangSplat~\cite{qin2024langsplat}
            & 8.87 & 21.78 & 3.04 & 11.39 & 11.27
            & 7.89 & \underline{38.10} & 3.39 & 13.79 & 15.79 \\
        OpenGaussian~\cite{wu2024opengaussian}
            & 3.96 & \underline{24.42} & 8.49 & 10.43 & 11.83
            & 2.63 & \textbf{39.68} & 10.17 & 20.69 & 18.29 \\
        Dr.Splat~\cite{drsplat25}
            & 6.21 & 7.25 & 2.90 & 6.76 & 5.78
            & 9.21 & 9.52 & 0.00 & 6.90 & 6.41 \\
        ReferSplat~\cite{he2025refersplat}
            & 28.81 & 14.14 & 35.54 & \textbf{23.44} & 25.48
            & \underline{39.47} & 17.46 & \underline{55.93} & \textbf{31.03} & 35.97 \\
        ReferSplat (w/ our loss)
            & \underline{33.06} & 16.82 & \underline{36.26} & 20.21 & \underline{26.59}
            & 38.16 & 25.40 & 54.24 & \underline{27.59} & \underline{36.35} \\
        ClusterSplat (Ours)
            & \textbf{35.27} & \textbf{26.18} & \textbf{41.39} & \underline{21.71} & \textbf{31.14}
            & \textbf{43.20} & 34.92 & \textbf{57.63} & \textbf{31.03} & \textbf{41.70} \\
        \bottomrule
    \end{tabular}
    }
\end{table}

%% file: appendix/B_6_efficiency.tex
\subsection{\pionerevision{Efficiency Analysis}}
\label{sec:efficiency_analysis}

\pionerevision{Table~\ref{tab:efficiency} reports preparation costs and online efficiency on an NVIDIA RTX A6000. ClusterSplat requires 125.3\,min of method-specific preparation per scene, 10.2\% more than ReferSplat. This cost is incurred once per scene, whereas online targeting is repeated for every expression. ClusterSplat takes 19.55\,ms per query, about 6.1 times faster than ReferSplat, while achieving higher grounding accuracy. The trade-off is therefore more preparation for lower latency during repeated interaction with a prepared scene.}

\pionerevision{Online targeting includes text encoding, cluster scoring, and target Gaussian selection; 2D rendering is measured separately. Offline costs average ten scenes and exclude shared RGB reconstruction and cache preparation. ClusterSplat's value sums separately timed instance-feature learning, cluster construction, and Stage~2 training with different scene representations.}

\input{tables/9_efficiency.tex}

%% file: tables/9_efficiency.tex
\begin{table}[H]
    \centering
    \caption{\pionerevision{Offline preparation costs and online efficiency on an NVIDIA RTX A6000. Offline costs cover method-specific stages, excluding RGB reconstruction and cache preparation. Accuracy is in percentages.} Bold and underlined values indicate the best and second-best distinct scores in each column.}
    \label{tab:efficiency}
    \renewcommand{\arraystretch}{1.2}
    \setlength{\tabcolsep}{5pt}

    \resizebox{\textwidth}{!}{
    \begin{tabular}{l c c c c c}
        \toprule
        \multirow{2}{*}{Method} & \pionerevision{Offline Preparation} & Online Targeting & Rendering & \multicolumn{2}{c}{ScanRefer Accuracy} \\
         & \pionerevision{Time (min/scene)$\downarrow$} & Time (ms/query)$\downarrow$ & FPS$\uparrow$ & 2D mIoU$\uparrow$ & 3D mIoU$\uparrow$ \\
        \midrule
        LangSplat & \pionerevision{\underline{21.6}} & 20.45 & 0.28 & 11.00 & 5.81 \\
        OpenGaussian & \pionerevision{25.8} & 43.74 & 13.10 & 18.48 & 1.06 \\
        Dr.Splat & \pionerevision{\textbf{2.0}} & \textbf{13.20} & \textbf{68.00} & 19.01 & \underline{9.48} \\
        ReferSplat & \pionerevision{113.7} & 119.71 & 1.13 & \underline{31.30} & 4.05 \\
        ClusterSplat (Ours) & \pionerevision{125.3} & \underline{19.55} & \underline{51.12} & \textbf{39.72} & \textbf{23.33} \\
        \bottomrule
    \end{tabular}
    }
\end{table}

%% file: appendix/B_7_qualitative.tex
\subsection{Additional Qualitative Results}
\label{subsec:qualitative}
\label{sec:app_qualitative}

We provide visual comparisons to examine model behavior under complex queries.

Figure~\ref{fig:sup_reflerf} complements the Ref-LERF quantitative comparison with fine-grained descriptions that include attributes, surface cues, and spatial relations. These examples test whether a method can localize targets described by detailed appearance cues, such as a green fruit-like object, writing on an object's surface, or a silver baking machine with a sleek exterior, as well as spatial prepositions such as the slices above the ramen noodles. Baseline masks often merge the target with nearby background, miss object parts, or spread support to distractor regions, whereas ClusterSplat keeps the predicted masks more concentrated on the referred objects across these examples.

\input{figures/6_reflerf.tex}

Figures~\ref{fig:sup_scanrefer_all_1} and~\ref{fig:sup_scanrefer_all_2} visualize ScanRefer qualitative results for all 10 evaluation scenes. These examples are intended to show qualitative coverage across the evaluation split, including attribute-based descriptions and spatial relations. The rows compare rendered 2D masks with explicit top-down 3D target selections, illustrating that ClusterSplat generally keeps support concentrated on the referred object while reducing distractor responses.

\input{figures/7_scanrefer.tex}
\input{figures/8_scanrefer.tex}

Figure~\ref{fig:sup_scannet_19class} provides an additional ScanNet v2 19-class visualization across all 10 evaluation scenes. The annotated point-cloud labels and Gaussian pseudo labels illustrate the evaluation label construction, while the method columns compare class-level scene labeling quality across 3DGS baselines.

\input{figures/9_scannet.tex}

%% file: figures/6_reflerf.tex
\begin{figure}[H]
    \centering
    \includegraphics[width=1.0\textwidth]{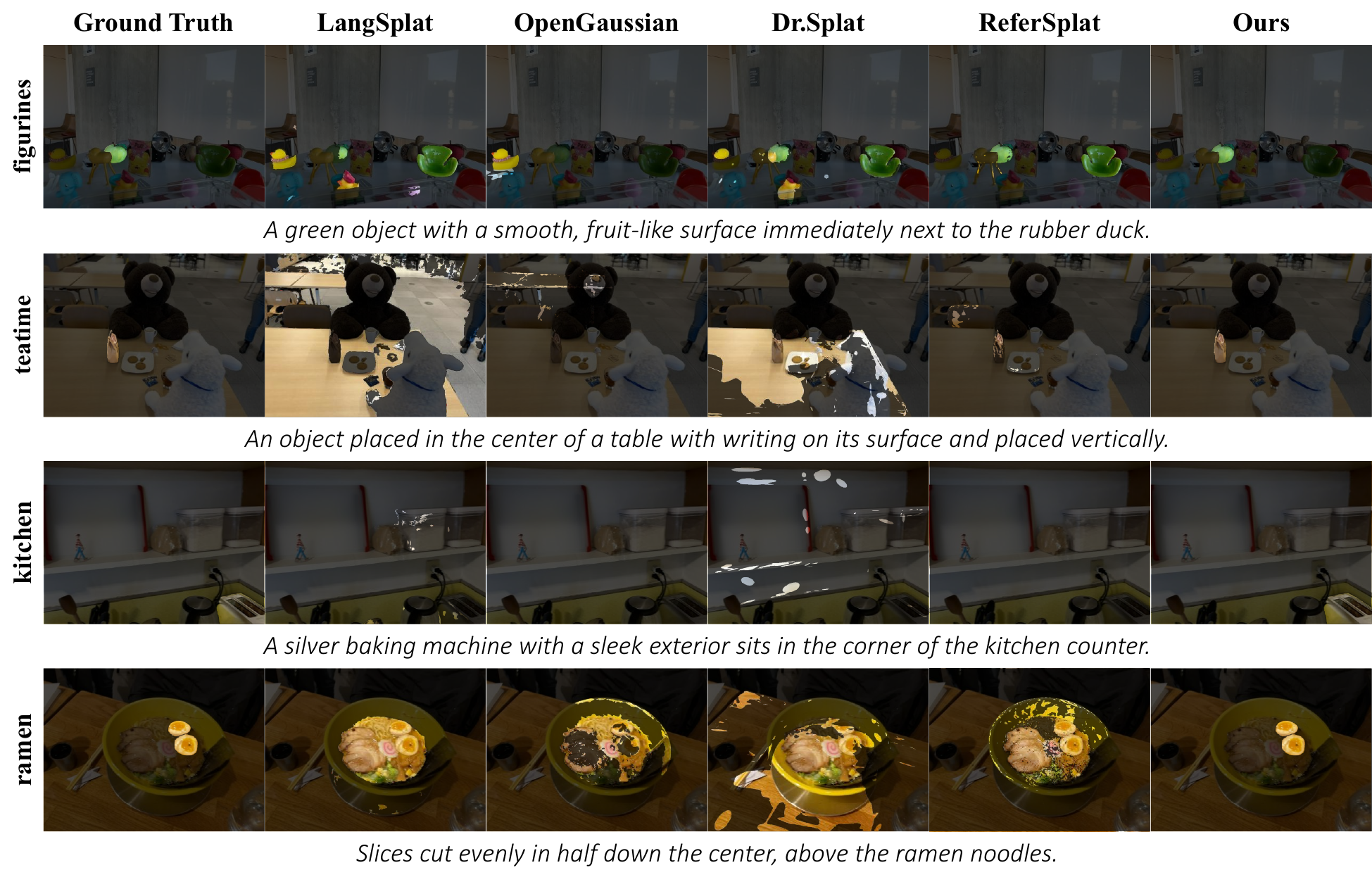}
    \caption{
        \textbf{Ref-LERF qualitative results.} Each row shows a fine-grained referring description with attributes, surface cues, or spatial relations. ClusterSplat produces compact masks that better follow the referred object across the shown examples.
    }
    \label{fig:sup_reflerf}
\end{figure}

%% file: figures/7_scanrefer.tex
\begin{figure}[p]
    \centering
    \includegraphics[height=0.88\textheight,keepaspectratio]{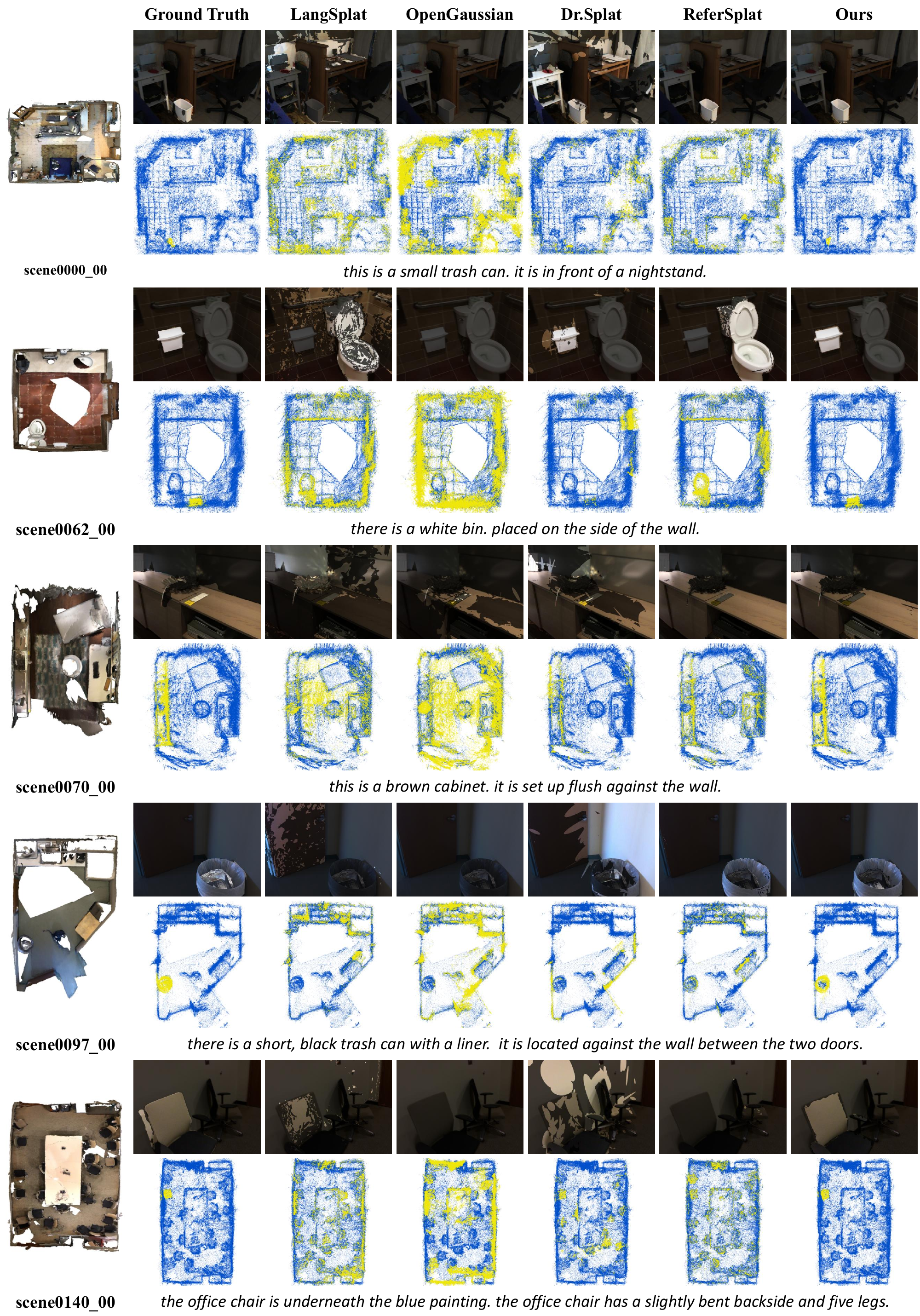}
    \caption{ScanRefer qualitative results on the first five evaluation scenes. Each row shows the ground-truth mask, rendered mask predictions, and top-down 3D target selections for a referring expression. The examples illustrate compact, object-aligned support with reduced distractor responses.}
    \label{fig:sup_scanrefer_all_1}
\end{figure}

%% file: figures/8_scanrefer.tex
\begin{figure}[p]
    \centering
    \includegraphics[height=0.88\textheight,keepaspectratio]{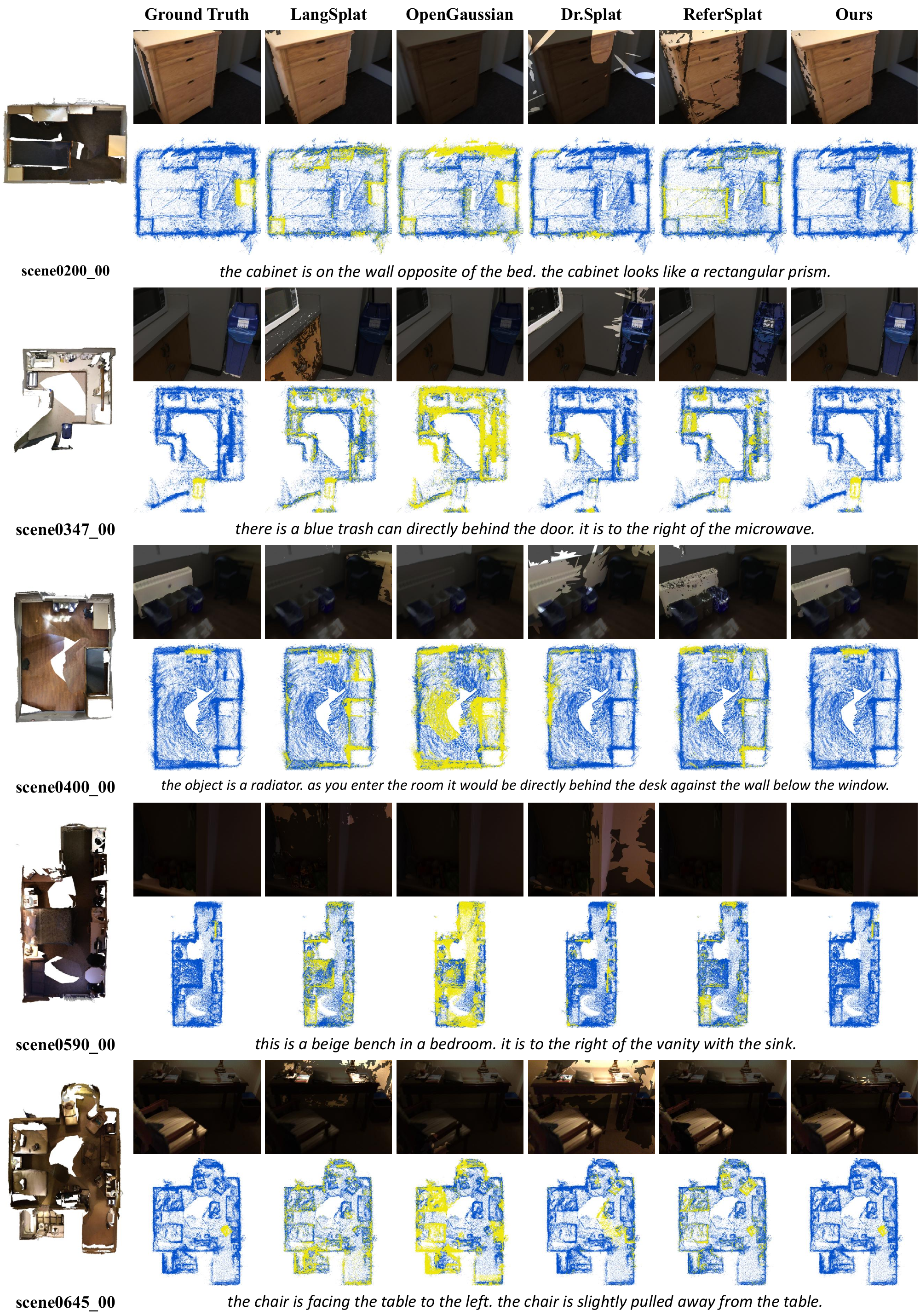}
    \caption{ScanRefer qualitative results on the remaining five evaluation scenes. The examples cover attribute-based and spatial referring expressions, showing consistent 2D mask concentration and top-down 3D target localization across the evaluation scenes.}
    \label{fig:sup_scanrefer_all_2}
\end{figure}

%% file: figures/9_scannet.tex
\begin{figure}[t]
    \centering
    \includegraphics[height=0.82\textheight,keepaspectratio]{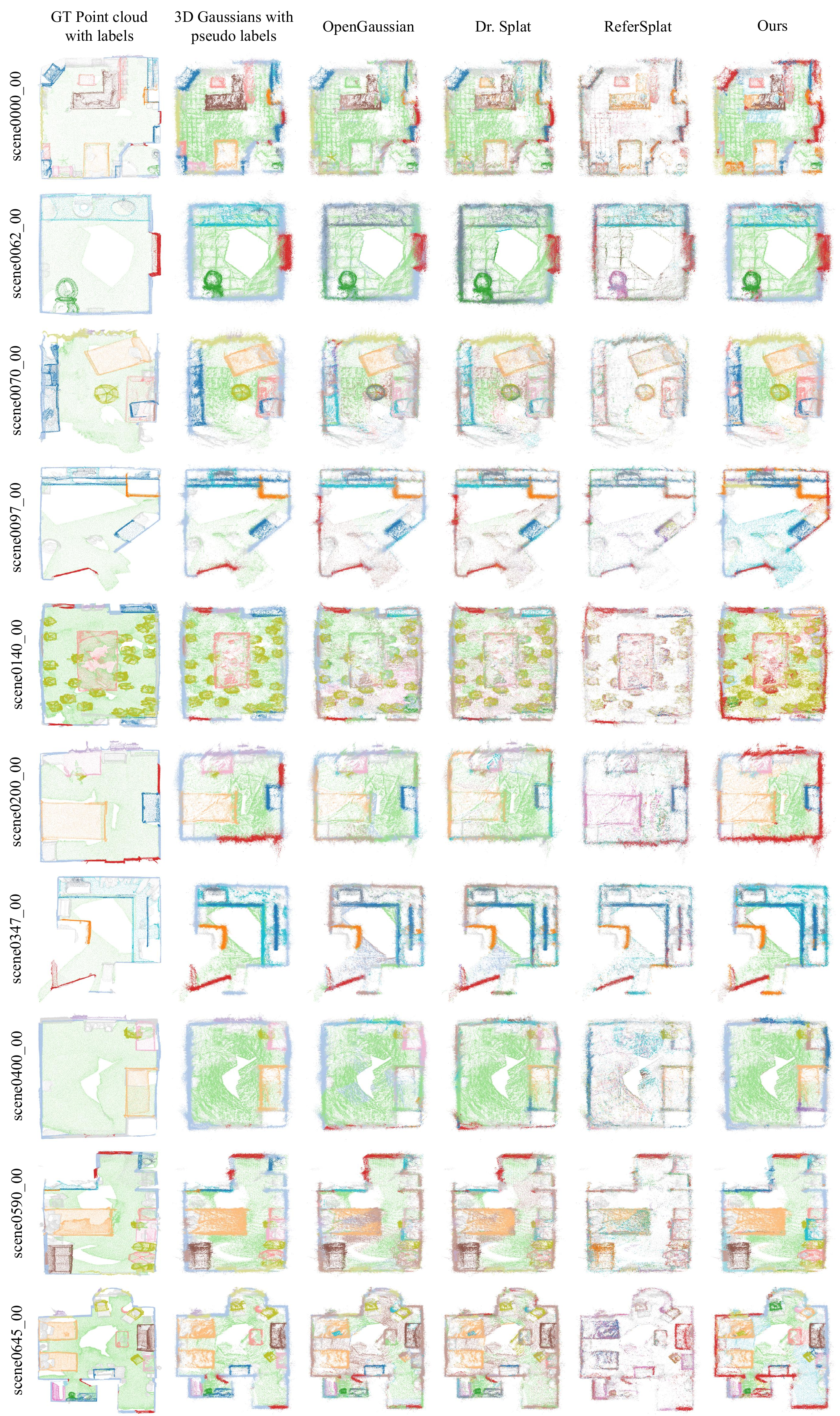}
    \vspace{-2mm}
    \caption{Top-down ScanNet v2 19-class visualization across all 10 evaluation scenes. The first two columns compare annotated ScanNet v2 point-cloud labels with Gaussian pseudo labels, followed by class-level predictions from 3DGS baselines and ClusterSplat.}
    \label{fig:sup_scannet_19class}
\end{figure}

%% file: checklist.tex
\section*{NeurIPS Paper Checklist}

\begin{enumerate}

\item {\bf Claims}
    \item[] Question: Do the main claims made in the abstract and introduction accurately reflect the paper's contributions and scope?
    \item[] Answer: \answerYes{} 
    \item[] Justification: The abstract and introduction state the 3D visual grounding setting, the cluster-level selection formulation, and the ScanNet v2/ScanRefer compatible evaluation protocol; these claims are supported by the empirical results in \pionerevision{Sections~\ref{sec:main_results} and~\ref{sec:cluster_construction_selection}, and Appendices~\ref{sec:efficiency_analysis} and~\ref{subsec:qualitative}.}

\item {\bf Limitations}
    \item[] Question: Does the paper discuss the limitations of the work performed by the authors?
    \item[] Answer: \answerYes{} 
    \item[] Justification: \pionerevision{Section~5 discusses cluster merging and fragmentation, single-target grounding, limited scene coverage, reliance on annotated supervision, and preparation costs for each new scene.}

\item {\bf Theory assumptions and proofs}
    \item[] Question: For each theoretical result, does the paper provide the full set of assumptions and a complete (and correct) proof?
    \item[] Answer: \answerNA{} 
    \item[] Justification: The paper does not present theoretical theorems or formal proof claims. The mathematical expressions define the model, clustering objective, and evaluation protocol.

    \item {\bf Experimental result reproducibility}
    \item[] Question: Does the paper fully disclose all the information needed to reproduce the main experimental results of the paper to the extent that it affects the main claims and/or conclusions of the paper (regardless of whether the code and data are provided or not)?
    \item[] Answer: \answerYes{} 
    \item[] Justification: Sections~3--4 and Appendix~A describe the model architecture, data preparation, held-out split, pseudo-label construction, baseline configurations, hyperparameters, and evaluation protocol. The supplementary material will include code and reproduction instructions.

\item {\bf Open access to data and code}
    \item[] Question: Does the paper provide open access to the data and code, with sufficient instructions to faithfully reproduce the main experimental results, as described in supplemental material?
    \item[] Answer: \answerYes{} 
    \item[] Justification: An anonymized supplementary code package will be submitted with training and evaluation scripts, configuration files, and data-preparation instructions. The paper uses public ScanNet v2, ScanRefer, and Ref-LERF resources, with preprocessing details described in Appendix~A.

\item {\bf Experimental setting/details}
    \item[] Question: Does the paper specify all the training and test details (e.g., data splits, hyperparameters, how they were chosen, type of optimizer) necessary to understand the results?
    \item[] Answer: \answerYes{} 
    \item[] Justification: Appendix~A.1 reports the software environment, optimizer, learning rate, training iterations, and architecture details. Appendix~A.4--A.6 describe the data split, supervision construction, evaluation protocol, and baseline settings.

\item {\bf Experiment statistical significance}
    \item[] Question: Does the paper report error bars suitably and correctly defined or other appropriate information about the statistical significance of the experiments?
    \item[] Answer: \answerYes{} 
    \item[] Justification: The experiments report quantitative benchmark statistics, including mIoU, mAcc, Top-1/Top-3 accuracy, ablation metrics, and efficiency measurements on fixed evaluation sets. All compared methods are evaluated under the same splits and metric definitions, so these reported statistics provide the basis for assessing the experimental differences.

\item {\bf Experiments compute resources}
    \item[] Question: For each experiment, does the paper provide sufficient information on the computer resources (type of compute workers, memory, time of execution) needed to reproduce the experiments?
    \item[] Answer: \answerYes{} 
    \item[] Justification: \pionerevision{Appendix~\ref{subsec:system_setup} specifies the RTX 5090 and RTX A6000 hardware and software environment used for the experiments. Appendix~\ref{sec:efficiency_analysis} and Table~\ref{tab:efficiency} report method-specific offline preparation time, online query latency, and rendering throughput.}
    
\item {\bf Code of ethics}
    \item[] Question: Does the research conducted in the paper conform, in every respect, with the NeurIPS Code of Ethics \url{https://neurips.cc/public/EthicsGuidelines}?
    \item[] Answer: \answerYes{} 
    \item[] Justification: The work uses established public research datasets and does not collect private data or conduct new human-subject studies. The submitted paper and supplementary material are anonymized for review.

\item {\bf Broader impacts}
    \item[] Question: Does the paper discuss both potential positive societal impacts and negative societal impacts of the work performed?
    \item[] Answer: \answerNA{} 
    \item[] Justification: The work is a technical study of 3D visual grounding on established public indoor-scene benchmarks. It does not collect new personal data, release a deployed user-facing system, or target safety-critical decision making, so we do not identify direct societal impacts specific to this method beyond the general research context of 3D scene understanding.
    
\item {\bf Safeguards}
    \item[] Question: Does the paper describe safeguards that have been put in place for responsible release of data or models that have a high risk for misuse (e.g., pre-trained language models, image generators, or scraped datasets)?
    \item[] Answer: \answerNA{} 
    \item[] Justification: The paper does not release a high-risk pre-trained language model, image generator, or scraped dataset. The supplementary release is code for reproducing the reported 3D visual grounding experiments.

\item {\bf Licenses for existing assets}
    \item[] Question: Are the creators or original owners of assets (e.g., code, data, models), used in the paper, properly credited and are the license and terms of use explicitly mentioned and properly respected?
    \item[] Answer: \answerYes{} 
    \item[] Justification: The paper cites the datasets, baselines, and models used in the experiments, including ScanNet v2, ScanRefer, Ref-LERF, CLIP, and the compared 3DGS baselines. The supplementary code package will include attribution, license notices where applicable, and instructions requiring users to follow the original dataset terms.

\item {\bf New assets}
    \item[] Question: Are new assets introduced in the paper well documented and is the documentation provided alongside the assets?
    \item[] Answer: \answerYes{} 
    \item[] Justification: The paper does not introduce a new dataset, but it will submit an anonymized supplementary code asset. The supplementary package will document installation, data preparation, training, and evaluation steps for reproducing the reported experiments.

\item {\bf Crowdsourcing and research with human subjects}
    \item[] Question: For crowdsourcing experiments and research with human subjects, does the paper include the full text of instructions given to participants and screenshots, if applicable, as well as details about compensation (if any)? 
    \item[] Answer: \answerNA{} 
    \item[] Justification: The work does not involve new crowdsourcing experiments or human-subject studies. It uses existing public datasets and annotations.

\item {\bf Institutional review board (IRB) approvals or equivalent for research with human subjects}
    \item[] Question: Does the paper describe potential risks incurred by study participants, whether such risks were disclosed to the subjects, and whether Institutional Review Board (IRB) approvals (or an equivalent approval/review based on the requirements of your country or institution) were obtained?
    \item[] Answer: \answerNA{} 
    \item[] Justification: The paper does not conduct new research with human subjects, so IRB approval or an equivalent review is not applicable.

\item {\bf Declaration of LLM usage}
    \item[] Question: Does the paper describe the usage of LLMs if it is an important, original, or non-standard component of the core methods in this research? Note that if the LLM is used only for writing, editing, or formatting purposes and does \emph{not} impact the core methodology, scientific rigor, or originality of the research, declaration is not required.
    \item[] Answer: \answerYes{} 
    \item[] Justification: Appendix~A.5 describes the use of GPT-4.1 as a VLM to generate pseudo-captions from training-frame instance masks. The appendix specifies the inputs provided to the model and states that official ScanRefer expressions, evaluation masks, and evaluation target IDs are not used in this generation process.

\end{enumerate}